\documentclass[10pt,twocolumn,letterpaper]{article}

\usepackage{wacv}              

\definecolor{wacvblue}{rgb}{0.21,0.49,0.74}
\usepackage[pagebackref,breaklinks,colorlinks,allcolors=wacvblue]{hyperref}

\usepackage[accsupp]{axessibility}
\usepackage{pifont}
\usepackage{multirow}
\usepackage{tabularx}

\usepackage{booktabs}
\usepackage{tabularx}
\usepackage[utf8]{inputenc}
\usepackage{textgreek}
\usepackage{multirow}
\usepackage{pifont}  
\usepackage{float}

\usepackage[most]{tcolorbox}
\usepackage{booktabs}
\usepackage{enumitem}
\usepackage[table]{xcolor}
\usepackage{amsmath}
\usepackage{amssymb}
\usepackage{subcaption}
\usepackage{pifont}
\usepackage{pifont}

\usepackage{graphicx}   
\usepackage{multirow}   
\usepackage{booktabs}   

\def\wacvPaperID{247} 
\def\confName{WACV}
\def\confYear{2027}

\title{Efficient Zero-Shot Video Anomaly Detection via Signal-Aware Adaptive Fusion (SAAF)}

\author{Khaled Waleed Dawoud \qquad Muhammad Haris Khan\\
Mohamed bin Zayed University of Artificial Intelligence\\
United Arab Emirates\\
{\tt\small \{khaled.dawoud, muhammad.haris\}@mbzuai.ac.ae}
}

\begin{document}
\maketitle
{\let\thefootnote\relax\footnotetext{\copyright~2027 IEEE. Personal use of this material is permitted. Permission from IEEE must be obtained for all other uses.}}
\begin{abstract}
Deploying Vision-Language Models for video anomaly detection
confronts a fundamental constraint. Their semantic demands computational budgets incompatible with the scale of real-world video streams. We present Signal-Aware Adaptive Fusion (SAAF), a framework that achieves efficient zero-shot anomaly detection through three key innovations. First, we introduce a dual-signal adaptive sampler operating in semantic embedding space: a delta signal captures sudden events via frame-to-frame cosine distance, while a drift signal detects static and gradual anomalies through deviation from rolling temporal context. Their OR-gated combination with adaptive EMA-based thresholding provides comprehensive anomaly coverage without training data. Second, we propose a threshold-gated cascaded architecture where lightweight frame-level scoring precedes expensive video-level temporal analysis, invoking it only when confidence warrants additional context. Third, we design a signal-aware weighted fusion mechanism that dynamically adapts score combination based on which signal triggered frame selection. 
Across four benchmarks, SAAF delivers competitive accuracy at substantially lower cost. Among training-free zero-shot methods, SAAF sets a new state of the art on XD-Violence (92.07\%), MSAD (86.70\%), and Shang (80.54\%) AUC-ROC, with XD and MSAD also surpassing supervised baselines.  
SAAF sustains 44.95, 48.61, and 41.06 FPS on XD, UCF, and MSAD, a 2.43×, 2.63×, and 2.22× throughput improvement over the prior best zero-shot method.
Code: \url{https://github.com/kdawoud91/SAAF}.
\end{abstract}
    
\section{Introduction}
\label{intro}

\begin{figure}[!t]
  \centering
  \includegraphics[width=\linewidth]{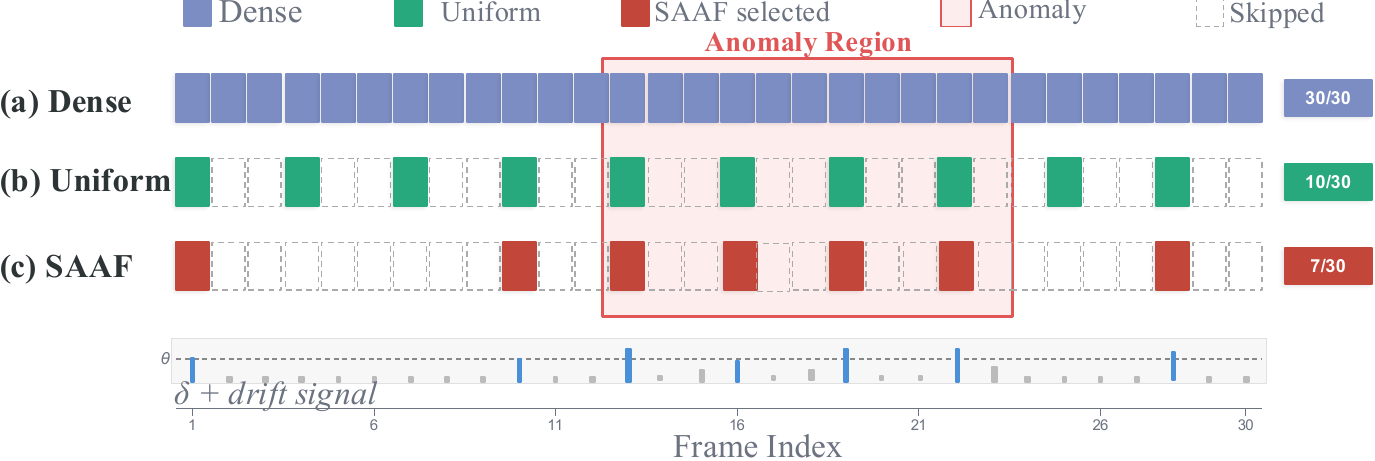}
    \caption{Comparison of sampling strategies. (a) Dense sampling. (b) Uniform sampling selects at fixed intervals. (c) SAAF adaptively selects within the same uniform stride pool, concentrating frames around anomalous regions (pink), avoiding redundant frames. 
    SAAF achieves efficient coverage of anomalous content while significantly reducing the number of processed frames.}
  \label{fig:teaser}
  \vspace{-0.5cm}
\end{figure}

Video anomaly detection (VAD) spans applications from public safety monitoring to industrial hazard oversight~\cite{dawoud2025fusedvision,sultani2018real,liu2018future}. Yet anomalies are inherently rare and unpredictable. Unlike object detection~\cite{redmon2016you,carion2020end}, where categories can be enumerated, what constitutes ``abnormal'' shifts across contexts and environments. This challenge has driven interest in zero-shot (ZS) approaches that detect anomalies without training on target categories. Vision-Language Models (VLMs) such as VideoLLaMA~\cite{zhang2023video}, Qwen-VL~\cite{bai2023qwen}, and GPT-4V~\cite{achiam2023gpt} offer promising capabilities from describing visual scenes, reasoning about behaviors, and assessing normalcy. In principle, a VLM could serve as a universal anomaly detector. However, in practice, computational cost presents a significant obstacle. One hour of 30 FPS video produces 108{,}000 frames, and even efficient VLMs such as~\cite{yang2025qwen3technicalreport} require hundreds of milliseconds per frame, rendering dense analysis impractical.

A more useful question is where in a video the VLM's capacity is
actually needed. Anomalies occupy a small fraction of most video
footage, and within an anomalous segment, frames vary in how much they
contribute to a correct decision. If computation could be steered toward the anomalous frames and the reasoning depth that each video actually requires, the VLM budget would shrink while concentrating on the frames that drive the decision. 
Existing pipelines offer no such mechanism, since frame selection, frame scoring, and temporal reasoning are each invoked on fixed intervals~\cite{zanella2024harnessing, lin2025unified,ye2025vera,wu2024weakly} with no mechanism to redirect effort based on what the video looks like.

\begin{figure*}[!t]
  \centering
\includegraphics[width=\textwidth, height=0.38\textheight]{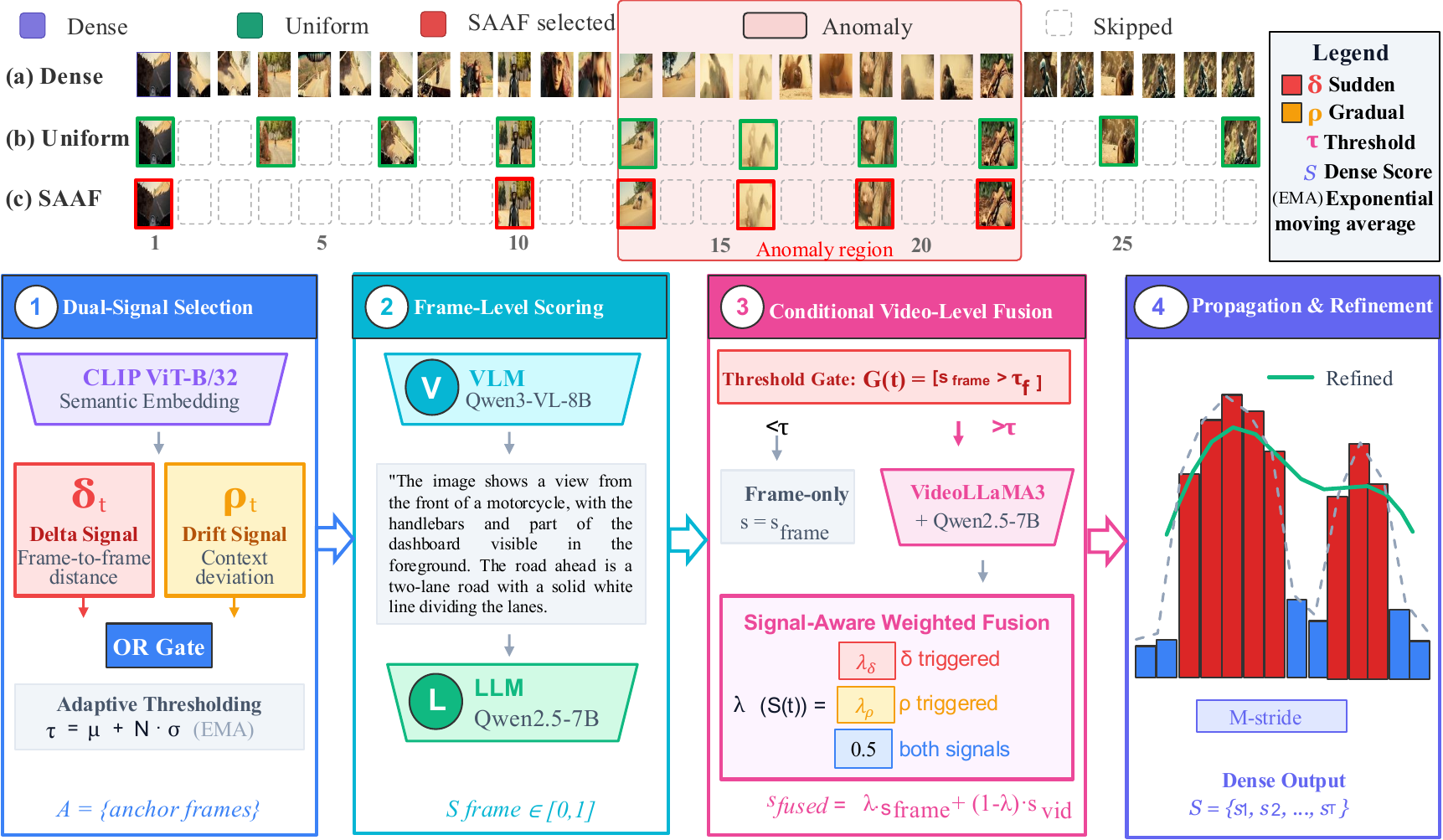}\caption{Overview of the SAAF pipeline. 
Stage 1 extracts delta (frame-to-frame change) and drift (context deviation) signals from CLIP embeddings via adaptive EMA-based thresholding. Stage 2 generates frame-level anomaly scores through VLM captioning and LLM reasoning. Stage 3 conditionally invokes video-level temporal analysis for high-scoring frames, combining frame and video scores via signal-aware weighted fusion. Stage 4 propagates and refines sparse anchor scores to produce dense frame-level predictions.}
  \label{fig:Methodology}
 \vspace{-0.5cm}
\end{figure*}

Notably, anomalies are not homogeneous in their temporal signature. \textit{Sudden events} such as explosions, shootings, and road accidents manifest as abrupt departures between adjacent
frames. \textit{Static or gradually evolving anomalies}, such as a slowly developing fire, produce frame-to-frame change yet remain abnormal relative to the surrounding context. The abnormality in the first case lies in what changes while in the second, in what persists. Uniform stride sampling is poorly suited to either regime. By allocating computation at fixed intervals, it processes redundant content while offering no mechanism to concentrate frames where either signature occurs. This motivates our central question: \textit{how
can adaptive sampling cover both abrupt and gradual anomalies while preserving efficiency on normal content?}

We propose a dual-signal adaptive sampler combining complementary signals. The \textit{delta signal} measures frame-to-frame change, capturing sudden events. The \textit{drift signal} measures deviation from a rolling temporal context, detecting divergence from recent patterns. Both operate on semantic embeddings, and support zero-shot operation through vision-language alignment, see Fig.~\ref{fig:teaser}.

Beyond frame selection, existing VLM-based systems typically run
both frame-level and video-level analysis unconditionally for
every sampled frame \cite{zanella2024harnessing}, regardless of
whether temporal context is actually needed. Our threshold-gated
cascaded architecture addresses this inefficiency. Among the
anchor frames retained by the dual-signal sampler, we first
compute frame-level scores and invoke video-level analysis only
on those whose frame-level score exceeds a confidence threshold.
This two-level gating, adaptive selection followed by score-based
cascading, sharply reduces the number of expensive video-level
VLM calls while focusing temporal reasoning where it is most
likely to add discriminative value.

For frames that trigger both analyses, naive score averaging ignores that different anomaly types benefit differently from each level. Our key insight is that the triggering signal itself reveals the anomaly type. Delta-triggered frames correspond to sudden events where frame-level analysis is most informative. Drift-triggered frames correspond to static or gradual anomalies where video-level context clarifies ambiguity. When both signals fire, equal weighting is applied. This yields our \textit{signal-aware weighted fusion} mechanism, which dynamically adjusts fusion weights based on which signal caused frame selection. We summarize our key contributions as follows:

\begin{itemize}
    \item \textbf{Dual-signal adaptive sampling.} We introduce complementary temporal signals in semantic embedding space: a delta signal capturing sudden transitions via frame-to-frame cosine distance, and a drift signal detecting static and gradual anomalies through deviation from rolling temporal context. Their OR-gated combination with adaptive EMA thresholding provides coverage of distinct anomaly types without training data.

    \item \textbf{Threshold-gated cascading.} Lightweight frame-level scoring gates expensive video-level analysis, invoking temporal reasoning only when confidence warrants it. 
    
    \item \textbf{Signal-aware weighted fusion.} The triggering signal informs score combination. Delta-triggered frames weight frame-level scores higher, drift-triggered frames weight video-level scores higher. This couples selection rationale to adaptive fusion strategy without additional parameters.

   \item \textbf{Efficiency and accuracy.} Among training-free zero-shot methods, SAAF sets a new SOTA on XD, MSAD, and Shang, with XD and MSAD also surpassing supervised baselines. SAAF sustains amortized throughputs of 44.95, 48.61, 41.06 and 33.55 FPS on XD, UCF, MSAD and Shang delivering 2.43$\times$, 2.63$\times$, 2.22$\times$ and 1.81$\times$ speedups over the prior best zero-shot method~\cite{lin2025unified}.
\end{itemize}

\section{Literature Review}
Traditional VAD methods, including reconstruction-based autoencoders~\cite{dawoud2025fusedvision, hasan2016learning,medel2016anomaly}, future-frame prediction~\cite{liu2018future,park2020learning,shen2022video}, and distance-based outlier detection~\cite{wang2019gods,micorek2024mulde}, operate as black boxes that produce anomaly scores without explaining what is anomalous or why. These approaches typically require training data and can be sensitive to distribution shifts across scenes and anomaly categories. 

Vision-Language Models (VLMs)~\cite{zhang2023video,alayrac2022flamingo,radford2021learning,li2023blip} and Large Language Models (LLMs)~\cite{grattafiori2024llama,achiam2023gpt,touvron2023llama} have enabled zero-shot anomaly detection with semantic interpretability. LAVAD~\cite{zanella2024harnessing} pioneered this direction with a training-free captioner-LLM-scorer pipeline. However, its uniform sampling is content-blind and it unconditionally invokes both frame-level and temporal-level analysis for every sampled frame, processing redundant normal content while failing to concentrate computation on anomalies.
Recent work explores multi-stage reasoning for zero-shot detection. Lin et al.~\cite{lin2025unified} propose a unified framework that chains temporal detection with spatial localization and textual understanding, using score-based gates to trigger additional reasoning for ambiguous predictions. However, their approach retains limitations: initial scoring operates on uniformly sampled frames, anomaly context is extracted after scoring rather than guiding frame selection, and reasoning refinement treats all anomaly types identically despite their differing analysis requirements.

\Cref{tab:sampling-comparison} summarises the sampling strategies adopted
by recent zero-shot and weakly-supervised VAD pipelines. The dominant
choice is a fixed stride, typically every sixteen
frames~\cite{zanella2024harnessing, lin2025unified, wu2024weakly}, with the
remaining methods substituting equally content-independent
heuristics, fixed segment lengths~\cite{ahn2026anyanomaly}, fixed-rate
sampling within sliding windows~\cite{ye2025vera}, or event-uniform
sampling after a separate detector~\cite{shao2025eventvad}.
Across the table, no method allows the video content to drive selection at the first stage which is where the largest computational savings are available, since every frame eliminated here removes one downstream VLM call.

More broadly, routing computation by content is a long-standing principle in efficient video understanding. AdaFrame~\cite{wu2019adaframe} learns a frame-selection policy, SCSampler~\cite{korbar2019scsampler} forwards only salient clips to a 3D backbone, AR-Net~\cite{meng2020ar} predicts per-frame input resolutions, FrameExit~\cite{ghodrati2021frameexit} exits early from intermediate features, and LiteEval~\cite{wu2019liteeval} gates between coarse and fine networks. SAAF is distinguished from prior work by two design choices. First, their saliency signal is \emph{learned} under closed-set supervision, a policy or gate is trained against action labels that define what counts as informative. Zero-shot VAD has no such labels, so the first-stage signal must be derived intrinsically rather than learned. Second, their notion of saliency is \emph{single-mode} a unified score of how much a frame contributes to a classification objective which suffices when the target event has a consistent temporal signature but fails when the same pipeline must surface both abrupt and persistent phenomena. We address both by replacing the selector with statistical deviation in a frozen semantic embedding space. The delta signal measures frame-to-frame change in CLIP space, 
while the drift signal measures deviation from a rolling temporal context. 
The result is a content-aware stage that requires no training data, properties that make the same selector applicable across traffic, industrial, and crowd scenarios without per-domain adaptation.

\begin{table}[h]
\centering
\small
\caption{Sampling strategies adopted by ZS \&
WS VAD methods.}
\label{tab:sampling-comparison}
\begin{tabular*}{\linewidth}{@{\extracolsep{\fill}}ll}
\toprule
\textbf{Method} & \textbf{Sampling Strategy} \\
\midrule
AnyAnomaly~\cite{ahn2026anyanomaly}      & Segment length 24, 4 frames per segment          \\
EventVAD~\cite{shao2025eventvad}         & Stride 16 per event (MLLM)     \\
Holmes-VAU~\cite{zhang2025holmes}        & Stride 16, then 16 frames via ATS                \\
VERA~\cite{ye2025vera}                   & 8 frames per 10\,s window                       \\
STPrompt~\cite{wu2024weakly}             & Stride 16 (UCF), stride 4 (Shang)                 \\
LAVAD~\cite{zanella2024harnessing}       & Stride 16                                        \\
Unified~\cite{lin2025unified}  & Stride 16                    \\
\midrule
\textbf{SAAF (Ours)} & Dual-signal $\delta\!\vee\!\rho$ on CLIP, EMA-thresholded  \\
\bottomrule
\end{tabular*}
\end{table}
\section{Methodology}
We propose SAAF, a zero-shot framework for efficient and robust video anomaly detection. Given a video $V = \{f_1, f_2, \dots, f_T\}$ consisting of $T$ frames, our objective is to produce a dense anomaly score sequence $S = \{s_1, s_2, \dots, s_T\}$, where each $s_t \in [0,1]$ represents the likelihood of anomalous activity at frame $t$.

The framework addresses three fundamental challenges: (i) identifying \emph{where} to allocate VLM computation, (ii) determining \emph{when} temporal reasoning is necessary, and (iii) deciding \emph{how} to fuse multi-level predictions. To this end, SAAF consists of four stages: (1) dual-signal adaptive frame selection, (2) frame-level anomaly scoring, (3) conditional video-level fusion, (4) score propagation and refinement. Figure~\ref{fig:Methodology} illustrates the overall pipeline.

\subsection{Dual-Signal Adaptive Frame Selection}

Applying VLM inference densely over all frames is computationally prohibitive and often redundant. Uniform sampling reduces cost but remains content-blind. We therefore aim to identify a sparse set of \emph{informative anchor frames} that are likely to contain anomalous content.

Formally, we seek an indicator function $\mathcal{I}: \{1, \dots, T\} \rightarrow \{0, 1\}$ such that:
\begin{equation}
\mathcal{I}(t) = 
\begin{cases}
1 & \text{if frame } f_t \text{ warrants VLM analysis} \\
0 & \text{otherwise}
\end{cases}
\end{equation}

A key motivating our design is that anomalies are not homogeneous: different anomaly types exhibit distinct temporal signatures. Sudden events manifest as abrupt visual transitions, whereas static or gradual anomalies evolve slowly or persist without noticeable frame-to-frame change. Capturing both requires complementary signals operating in semantic embedding space.

\subsubsection{\textbf{Semantic Frame Embedding}}

For each frame $f_t$, we extract a semantic embedding using a pretrained visual encoder $\phi$:
\begin{equation}
e_t = \frac{\phi(f_t)}{\|\phi(f_t)\|_2} \in \mathbb{R}^d
\end{equation}

By default, we adopt CLIP ViT-B/32~\cite{radford2021learning} as our encoder $\phi$ due to its strong alignment between visual and semantic concepts. The $\ell_2$-normalization ensures $\|e_t\|_2 = 1$, enabling cosine similarity as a unified distance metric:
\begin{equation}
\text{sim}(e_i, e_j) = e_i \cdot e_j = \cos(\theta_{ij})
\end{equation}
where $\theta_{ij}$ is the angle between embeddings. 

\subsubsection{\textbf{Delta Signal: Instantaneous Semantic Change}}

We define the delta signal as the cosine distance between consecutive embeddings:
\begin{equation}
\delta_t = 1 - e_t \cdot e_{t-1} = 1 - \cos(\theta_{t,t-1})
\end{equation}

This signal captures abrupt semantic transitions, with $\delta_t \in [0, 2]$ where $\delta_t = 0$ indicates identical frames and $\delta_t = 2$ indicates maximally dissimilar frames. In practice, consecutive video frames yield $\delta_t \in [0, 1]$ under normal conditions. A spike in $\delta_t$ indicates rapid change in scene composition or activity. A frame is selected when $\delta_t$ exceeds an adaptive threshold $\tau_{\delta,t}$:
\begin{equation}
\mathcal{I}_\delta(t) = \mathbb{1}[\delta_t > \tau_{\delta,t}]
\end{equation}
where $\mathbb{1}[\cdot]$ is the indicator function and $\tau_{\delta,t}$ is computed from running statistics as described in Section~\ref{sec:adaptive_threshold}.

However, delta alone is insufficient. Anomalies that persist without motion produce low $\delta_t$ despite being semantically abnormal.

\subsubsection{\textbf{Drift Signal: Deviation from Temporal Context}}

To capture persistent and gradually evolving anomalies, we introduce the \textit{drift} signal. We maintain a rolling temporal anchor over the preceding $W$ frames:
\begin{equation}
a_t = \frac{1}{W} \sum_{i=t-W}^{t-1} e_i
\end{equation}

The normalized anchor and drift distance are defined as:
\begin{equation}
\hat{a}_t = \frac{a_t}{\|a_t\|_2}, \quad \rho_t = 1 - \hat{a}_t \cdot e_t
\end{equation}

Unlike delta, which captures instantaneous change, drift measures cumulative deviation from recent temporal context. A frame is selected when drift exceeds its adaptive threshold:
\begin{equation}
\mathcal{I}_\rho(t) = \mathbb{1}[\rho_t > \tau_{\rho,t}]
\end{equation}

\subsubsection{\textbf{Adaptive Thresholding}}
\label{sec:adaptive_threshold}

Fixed thresholds are brittle across diverse scenes with varying baseline dynamics. We therefore employ adaptive thresholds based on running statistics computed via exponential moving average (EMA).

For each signal $x \in \{\delta, \rho\}$, we maintain online estimates of the first moment (mean) $\mu_x^{(t)}$ and second moment (mean of squares) $\nu_x^{(t)}$:
\begin{align}
\mu_x^{(t)} &= (1 - \alpha)\mu_x^{(t-1)} + \alpha x_t \\
\nu_x^{(t)} &= (1 - \alpha)\nu_x^{(t-1)} + \alpha x_t^2
\end{align}
where $\alpha \in (0,1)$ controls the adaptation rate and $x_t$ denotes the value at frame $t$. The running variance is derived as:
\begin{equation}
(\sigma_x^{(t)})^2 = \nu_x^{(t)} - (\mu_x^{(t)})^2
\end{equation}

The adaptive threshold follows a sigma-rule formulation:
\begin{equation}
\tau_{x,t} = \mu_x^{(t)} + N \cdot \sigma_x^{(t)}
\end{equation}
where $N$ controls detection sensitivity. Higher $N$ raises the threshold and reduces the selection rate. We treat $N$ as a sensitivity parameter (see Supp.~A.2 for a sweep across $N \in \{0.5, 1.0, 1.5, 2.0\}$).

\textbf{Statistic Freeze.}
When a frame is selected, we suspend updates to the running statistics:
\begin{equation}
(\mu_x^{(t)}, \sigma_x^{(t)}) = 
\begin{cases}
(\mu_x^{(t-1)}, \sigma_x^{(t-1)}) & \text{if } \mathcal{I}(t) = 1 \\
\text{EMA update} & \text{otherwise}
\end{cases}
\end{equation}

This mechanism prevents anomalous observations from corrupting baseline statistics.

\subsubsection{\textbf{OR-Gated Selection and Signal Attribution}}
\label{sec:or_gated}

The final selection indicator combines both signals via logical disjunction:
\begin{equation}
\mathcal{I}(t) = \mathcal{I}_\delta(t) \lor \mathcal{I}_\rho(t) = \mathbb{1}[\delta_t > \tau_{\delta,t}] \lor \mathbb{1}[\rho_t > \tau_{\rho,t}]
\end{equation}

This OR-gated design ensures complementary coverage across anomaly types. We additionally define a signal attribution function $\mathcal{S}: \mathcal{A} \rightarrow \{\texttt{delta}, \texttt{drift}, \texttt{both}\}$:
\begin{equation}
\mathcal{S}(t) = 
\begin{cases}
\texttt{delta} & \text{if } \mathcal{I}_\delta(t) = 1 \land \mathcal{I}_\rho(t) = 0 \\
\texttt{drift} & \text{if } \mathcal{I}_\delta(t) = 0 \land \mathcal{I}_\rho(t) = 1 \\
\texttt{both} & \text{if } \mathcal{I}_\delta(t) = 1 \land \mathcal{I}_\rho(t) = 1
\end{cases}
\end{equation}

Critically, $\delta$ and $\rho$ are not anomaly categories but orthogonal modes of temporal change. An anomalous event might exhibit an abrupt departure from the immediate past, a sustained deviation from the rolling context, or both. The OR-gate covers all three regimes. 

\subsection{Frame-Level Anomaly Scoring}
\label{sec:frame_scoring}

Selected frames undergo semantic analysis via a two-stage VLM-LLM pipeline. For each selected frame $f_t$ where $\mathcal{I}(t) = 1$, we compute an anomaly score through visual captioning followed by language-based reasoning.

\subsubsection{\textbf{Visual Captioning}}
We employ a vision-language model $\mathcal{V}$ to generate a natural language description:
\begin{equation}
c_t = \mathcal{V}(f_t, p_{\text{cap}})
\end{equation}
where $p_{\text{cap}}$ is the captioning prompt.  The prompt instructs the model to describe the scene with emphasis on actions, and context (see Supplementary Section C for exact prompts). 

\subsubsection{Language-Based Anomaly Reasoning}
The generated caption is evaluated by a large language model $\mathcal{L}$ to produce an anomaly score:
\begin{equation}
s_t^{\text{frame}} = \mathcal{L}(c_t, p_{\text{score}}) \in [0, 1]
\end{equation}
where $p_{\text{score}}$ is the scoring prompt. The prompt instructs the model to assess whether the scene description depicts normal or anomalous behavior and output a score in $[0,1]$ (see Supplementary Section C). 

The complete frame-level scoring function is:
\begin{equation}
s_t^{\text{frame}} = 
\begin{cases}
\mathcal{L}(\mathcal{V}(f_t, p_{\text{cap}}), p_{\text{score}}) & \text{if } \mathcal{I}(t) = 1 \\
0 & \text{otherwise}
\end{cases}
\end{equation}

\subsection{Conditional Video-Level Fusion}
\label{sec:fusion}

While frame-level analysis captures visually salient anomalies, certain events require temporal context for disambiguation. We introduce conditional video-level reasoning that incorporates temporal context when warranted.

\subsubsection{\textbf{Threshold-Gated Temporal Reasoning}}

We apply video-level analysis selectively based on frame-level confidence. Let $\tau_f$ denote the fusion threshold. Video-level reasoning is triggered when:
\begin{equation}
\mathcal{G}(t) = \mathbb{1}[s_t^{\text{frame}} > \tau_f]
\end{equation}

For frames where $\mathcal{G}(t) = 1$, we extract a temporal clip $C_t = \{f_{t-k}, \dots, f_t, \dots, f_{t+k}\}$ 
and process it through a video-language model $\mathcal{V}_{\text{vid}}$ followed by a language model:
\begin{equation}
s_t^{\text{vid}} = \mathcal{L}(\mathcal{V}_{\text{vid}}(C_t, p_{\text{vid}}), p_{\text{score}})
\end{equation}

This achieves reduction in computational cost while focusing temporal reasoning on high-confidence candidates.

\subsubsection{\textbf{Signal-Aware Weighted Fusion}}
For frames receiving both scores, we adapt fusion weights based on the triggering signal $\mathcal{S}(t)$. We define the signal-dependent weight function $\lambda$ mapping $\{\texttt{delta}, \texttt{drift}, \texttt{both}\} \to [0, 1]$:
\begin{equation}
\lambda(\mathcal{S}(t)) = 
\begin{cases}
\lambda_\delta & \text{if } \mathcal{S}(t) = \texttt{delta} \\
\lambda_\rho & \text{if } \mathcal{S}(t) = \texttt{drift} \\
\lambda_{\text{both}} & \text{if } \mathcal{S}(t) = \texttt{both}
\end{cases}
\end{equation}
where $\lambda_\delta, \lambda_\rho, \lambda_{\text{both}} \in [0,1]$ are fusion weights.

The fused anomaly score is computed as a combination:
\begin{equation}
s_t^{\text{fused}} = \lambda(\mathcal{S}(t)) \cdot s_t^{\text{frame}} + (1 - \lambda(\mathcal{S}(t))) \cdot s_t^{\text{vid}}
\end{equation}

The scoring function integrating all conditions is:
\begin{equation}
s_t = 
\begin{cases}
s_t^{\text{fused}} & \text{if } \mathcal{I}(t) = 1 \land \mathcal{G}(t) = 1 \\
s_t^{\text{frame}} & \text{if } \mathcal{I}(t) = 1 \land \mathcal{G}(t) = 0 \\
0 & \text{if } \mathcal{I}(t) = 0
\end{cases}
\end{equation}

\begin{figure*}[t]
    \centering
    \begin{subfigure}[b]{0.46\textwidth}
        \centering
        \includegraphics[width=\textwidth]{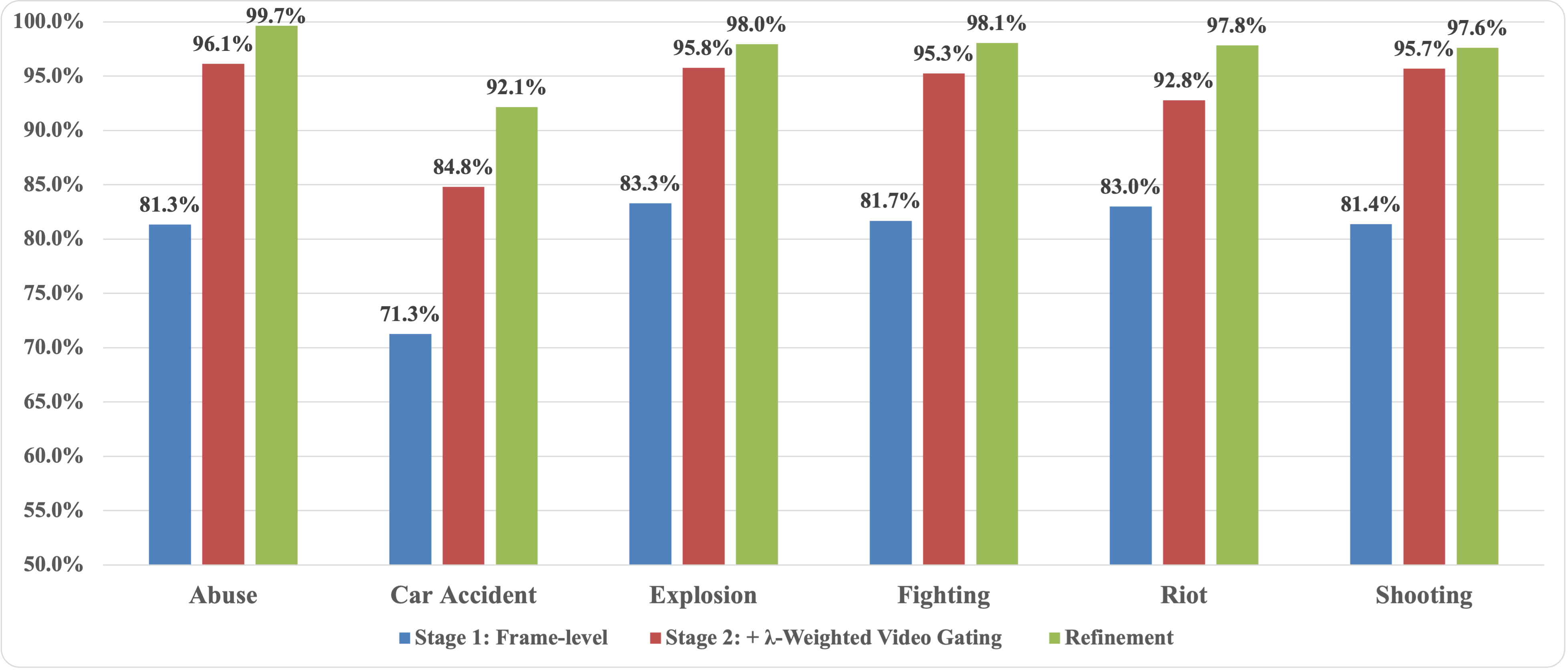}
        \caption{XD (6 anomaly categories)}
        \label{fig:breakdown-xd}
    \end{subfigure}
    \hfill
    \begin{subfigure}[b]{0.52\textwidth}
        \centering
        \includegraphics[width=\textwidth]{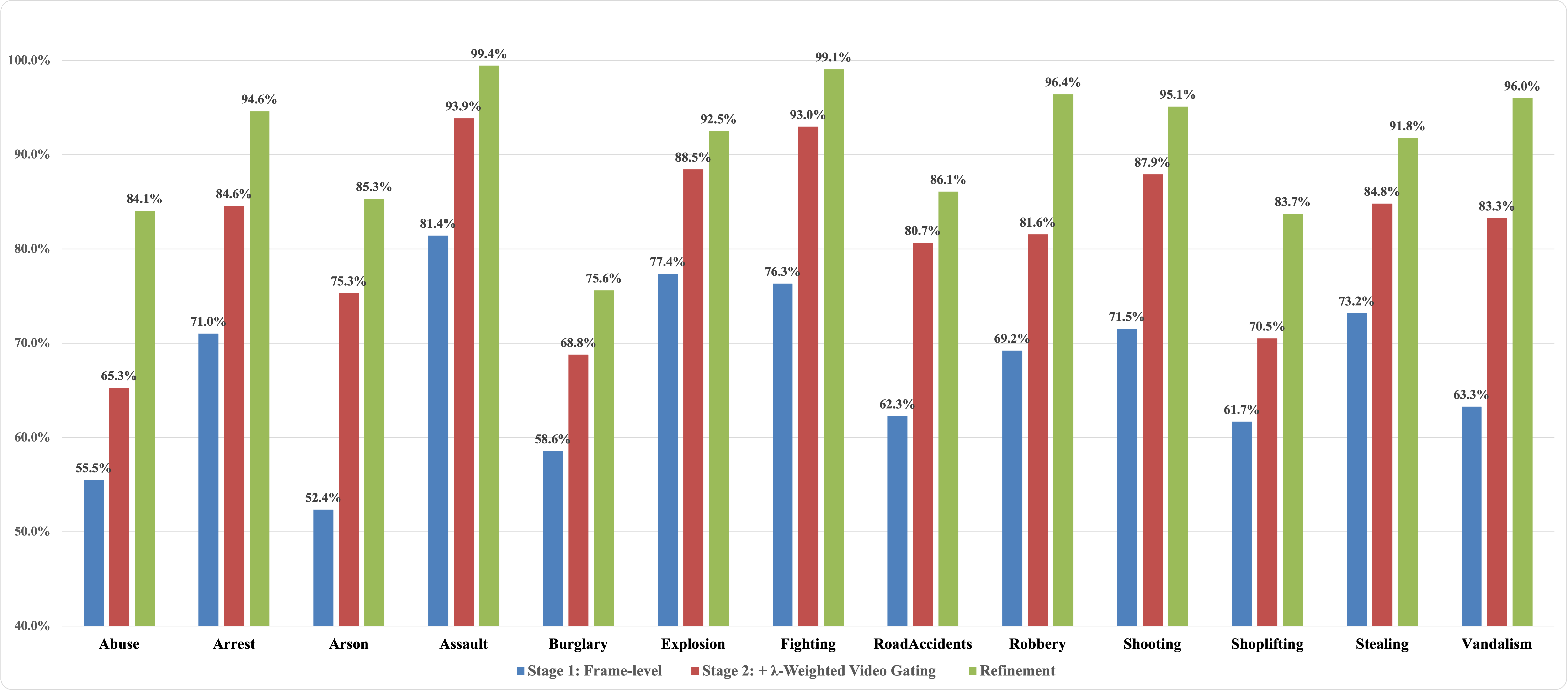}
        \caption{UCF-Crime (13 anomaly categories)}
        \label{fig:breakdown-ucf}
    \end{subfigure}

\caption{Per-category AUC-ROC across SAAF pipeline stages. Stage~2 (blue): frame-level scoring. Stage~3 (red): $\lambda$-weighted video-level gating. Stage~4 (green): refinement. All categories benefit from multi-stage fusion.}
\label{fig:per-category-breakdown}
\vspace{-0.5cm}
\end{figure*}

\subsection{Score Propagation}

Since VLM inference is performed only at selected frames, we propagate scores temporally to obtain dense predictions. Let $M$ denote the maximum propagation stride and $\mathrm{next}(t)$ the next anchor after $t \in \mathcal{A}$. For each anchor with computed score $s_t$, we define the propagation operator:
\begin{equation}
\begin{aligned}
\mathcal{P}(s_t) &= \{(t + i, s_t) : i \in \{0, 1, \ldots, \ell_t - 1\}\}, \\
\ell_t &= \min(\mathrm{next}(t) - t,\, M)
\end{aligned}
\end{equation}

The dense score sequence is constructed as:
\begin{equation}
\mathcal{S} = \bigcup_{t \in \mathcal{A}} \mathcal{P}(s_t)
\end{equation}

Frames outside any propagation window receive a default fill value $s_d$:
\begin{equation}
s_j =
\begin{cases}
s_t & \text{if } \exists\, t \in \mathcal{A} : j \in [t,\, t + \ell_t) \\
s_d & \text{otherwise}
\end{cases}
\end{equation}

\begin{table}[!t]
\centering
\caption{VAD Performance comparison across XD, UCF, MSAD and Shang. \textcolor{green!70!black}{\ding{51}} / \textcolor{red}{\ding{55}} indicate whether a method is zero-shot and training-free in terms of model parameters. Our SAAF results are mean $\pm$ std over 3 runs.}
\label{tab:vad_comparison}
\scriptsize
\setlength{\tabcolsep}{3pt}
\renewcommand{\arraystretch}{1.05}
\resizebox{\columnwidth}{!}{%
\begin{tabular}{lcccccc}
\toprule
\multirow{2}{*}{Method} & \multirow{2}{*}{Zero-shot} & \multirow{2}{*}{Training-free} & XD~\cite{wu2020not} & UCF~\cite{sultani2018real} & MSAD~\cite{zhu2024advancing} & Shang~\cite{luo2017revisit} \\
\cmidrule(lr){4-4} \cmidrule(lr){5-5} \cmidrule(lr){6-6} \cmidrule(lr){7-7}
 & & & AUC(\%) & AUC(\%) & AUC(\%) & AUC(\%) \\
\midrule
Sultani et al.~\cite{sultani2018real} & \textcolor{red}{\ding{55}} & \textcolor{red}{\ding{55}} & -- & 77.92 & -- & -- \\
FusedVision~\cite{dawoud2025fusedvision} & \textcolor{red}{\ding{55}} & \textcolor{red}{\ding{55}} & -- & -- & -- & \underline{83.58} \\
RTFM~\cite{tian2021weakly} & \textcolor{red}{\ding{55}} & \textcolor{red}{\ding{55}} & \underline{61.56} & -- & \textbf{85.67} & -- \\
GODS~\cite{wang2019gods} & \textcolor{red}{\ding{55}} & \textcolor{red}{\ding{55}} & -- & 70.46 & -- & -- \\
MGFN~\cite{chen2023mgfn} & \textcolor{red}{\ding{55}} & \textcolor{red}{\ding{55}} & -- & 86.98 & \underline{85.0} & -- \\
STPrompt~\cite{wu2024weakly} & \textcolor{red}{\ding{55}} & \textcolor{red}{\ding{55}} & -- & \underline{88.08} & -- & -- \\
OVVAD~\cite{wu2024open} & \textcolor{red}{\ding{55}} & \textcolor{red}{\ding{55}} & -- & 86.40 & -- & -- \\
Holmes-VAU~\cite{zhang2025holmes} & \textcolor{red}{\ding{55}} & \textcolor{red}{\ding{55}} & -- & \textbf{88.96} & -- & -- \\
MULDE~\cite{micorek2024mulde} & \textcolor{red}{\ding{55}} & \textcolor{red}{\ding{55}} & -- & - & -- & 81.40 \\
UR-DMU~\cite{zhou2023dual} & \textcolor{red}{\ding{55}} & \textcolor{red}{\ding{55}} & -- & -- & 74.30 & -- \\
AnomalyRuler~\cite{yang2024follow} & \textcolor{red}{\ding{55}} & \textcolor{green!70!black}{\ding{51}} & -- & -- & -- & \textbf{85.20} \\
VERA~\cite{ye2025vera} & \textcolor{red}{\ding{55}} & \textcolor{green!70!black}{\ding{51}} & \textbf{88.26} & 86.55 & -- & -- \\
\midrule
LLAVA-1.5~\cite{liu2024improved} & \textcolor{green!70!black}{\ding{51}} & \textcolor{green!70!black}{\ding{51}} & 79.62 & 72.84 & -- & -- \\
Eventvad~\cite{shao2025eventvad} & \textcolor{green!70!black}{\ding{51}} & \textcolor{green!70!black}{\ding{51}} & 87.51 & 82.30 & -- & -- \\
AnyAnomaly~\cite{ahn2026anyanomaly} & \textcolor{green!70!black}{\ding{51}} & \textcolor{green!70!black}{\ding{51}} & -- & 80.70 & -- & \underline{79.70} \\
MCANet~\cite{dev2024mcanet} & \textcolor{green!70!black}{\ding{51}} & \textcolor{green!70!black}{\ding{51}} & 87.43 & 82.47 & -- & -- \\
LAVAD~\cite{zanella2024harnessing} & \textcolor{green!70!black}{\ding{51}} & \textcolor{green!70!black}{\ding{51}} & 85.36 & 80.28 & -- & -- \\
Unified-reasoning~\cite{lin2025unified} & \textcolor{green!70!black}{\ding{51}} & \textcolor{green!70!black}{\ding{51}} & \underline{91.34} & \textbf{84.28} & \underline{85.90} & -- \\
\midrule
Ours (adaptive) & \textcolor{green!70!black}{\ding{51}} & \textcolor{green!70!black}{\ding{51}} & \textbf{92.07}$_{\pm 0.02}$ & \underline{82.59}$_{\pm 0.05}$ & \textbf{86.70}$_{\pm 0.05}$ & \textbf{80.54}$_{\pm 0.03}$ \\
\bottomrule
\end{tabular}%
}
\vspace{-0.5cm}
\end{table}

\section{Experiments}

\subsection{Implementation Details}
\textbf{Frame Sampling and Embedding Extraction.} We apply stride-16 sampling for signal computation, reducing the candidate pool to approximately 6.25\% of original frames. Each sampled frame is encoded via CLIP ViT-B/32, yielding 512-dimensional $\ell_2$-normalized embeddings. The dual-signal adaptive sampler operates on these embeddings with EMA-based adaptive thresholding (decay factor $\alpha = 0.05$), selecting frames whose signal values exceed $N = 1$ standard deviation above the running mean for VLM inference. A warmup period of 10 frames initializes the running statistics before detection begins. The rolling window for drift computation spans $W = 20$ frames. To preserve sensitivity during anomaly bursts, EMA statistics are frozen whenever a frame is selected, preventing anomalous observations from corrupting the baseline distribution.
\textbf{Frame-Level Scoring.} Selected anchor frames undergo visual captioning via Qwen3-VL-8B-Instruct~\cite{yang2025qwen3technicalreport}, followed by anomaly reasoning through Qwen2.5-7B-Instruct to produce scores in $[0, 1]$.
\textbf{Video-Level Fusion.} For frames whose frame-level score exceeds the threshold $\tau_f = 0.4$, we extract a temporal clip of 10 frames centered at the anchor and process it through VideoLLaMA3-7B~\cite{zhang2025videollama}. The resulting video-level caption is scored by Qwen2.5-7B-Instruct using the same anomaly prompt. Frame-level and video-level scores are combined via signal-aware weighted fusion as described in Section~\ref{sec:fusion}, with fusion weights set to $\lambda_\delta = 0.6$, $\lambda_\rho = 0.4$, and $\lambda_{\text{both}} = 0.5$. 
\noindent\textbf{Score Propagation and Refinement.} Anchor scores are propagated forward with a maximum stride $M = 64$; each anchor's score fills subsequent frames until either the next anchor or $M$ frames, whichever comes first. 
Frames where neither $\delta$ (instantaneous change) nor $\rho$ (contextual deviation) exceeded their respective adaptive thresholds remain unselected and receive a default score of $s_d = 0.3$, encoding a soft normality prior. 
All three defaults were fixed a priori. $N = 1 $ balances selection rate and AUC-ROC (Supp. A.2), sd = 0.3 is a conservative normality prior, τf = 0.4 is a low threshold chosen to preserve recall.
Following~\cite{lin2025unified}, We then apply Gaussian smoothing with $\sigma = 100$.
We then apply Gaussian smoothing on the dense per-frame score sequence with ($\approx 3.3$\,s at 30\,fps).

\subsection{Results \& Discussion}
Table~\ref{tab:vad_comparison} compares SAAF against supervised and zero-shot methods across four benchmarks.
\textbf{XD.} SAAF reaches 92.07\% AUC-ROC, surpassing the previous best zero-shot result of 91.34\% from Unified~\cite{lin2025unified} while processing 30.31\% fewer frames.
\textbf{UCF.} SAAF reaches 82.59\% AUC-ROC, outperforming LAVAD~\cite{zanella2024harnessing}, AnyAnomaly~\cite{ahn2026anyanomaly}, and Eventvad~\cite{shao2025eventvad} (80.28--82.30\%) and remaining competitive with Unified~\cite{lin2025unified} (84.28\%) despite processing 23.48\% fewer frames. Supervised methods Holmes-VAU~\cite{zhang2025holmes} and STPrompt~\cite{wu2024weakly} score higher (88.96, 88.08\%), but benefit from target-domain training and uniform processing.
\textbf{MSAD.} SAAF reaches 86.70\% AUC-ROC, the highest among supervised (RTFM~\cite{tian2021weakly}: 85.67\%, MGFN: 85.0\%) and zero-shot (Unified: 85.90\%) methods, with 34.87\% fewer frames processed.
\textbf{Shang.} SAAF reaches 80.54\% AUC-ROC competitive with supervised MULDE~\cite{micorek2024mulde} (81.40\%) and FusedVision~\cite{dawoud2025fusedvision} (83.58\%). Shang anomalies are brief, limiting the temporal resolution available to $\delta$ and $\rho$ under stride-16 sampling. 

\textbf{Two further analyses support these results}. The per-category breakdown (Fig.~\ref{fig:per-category-breakdown}) confirms the dual-signal taxonomy. Abrupt events (fighting, explosions, shootings) score well at the frame level, while gradual ones (abuse, robbery) gain substantially from video-level fusion. Smoothing applied directly to Stage~2 (Table 7) leaves a consistent margin to the full pipeline across benchmarks, showing Stage~3's semantic evidence drives the gain.

\begin{figure}[!t]
  \centering
  \includegraphics[width=.5\textwidth, height=6cm]{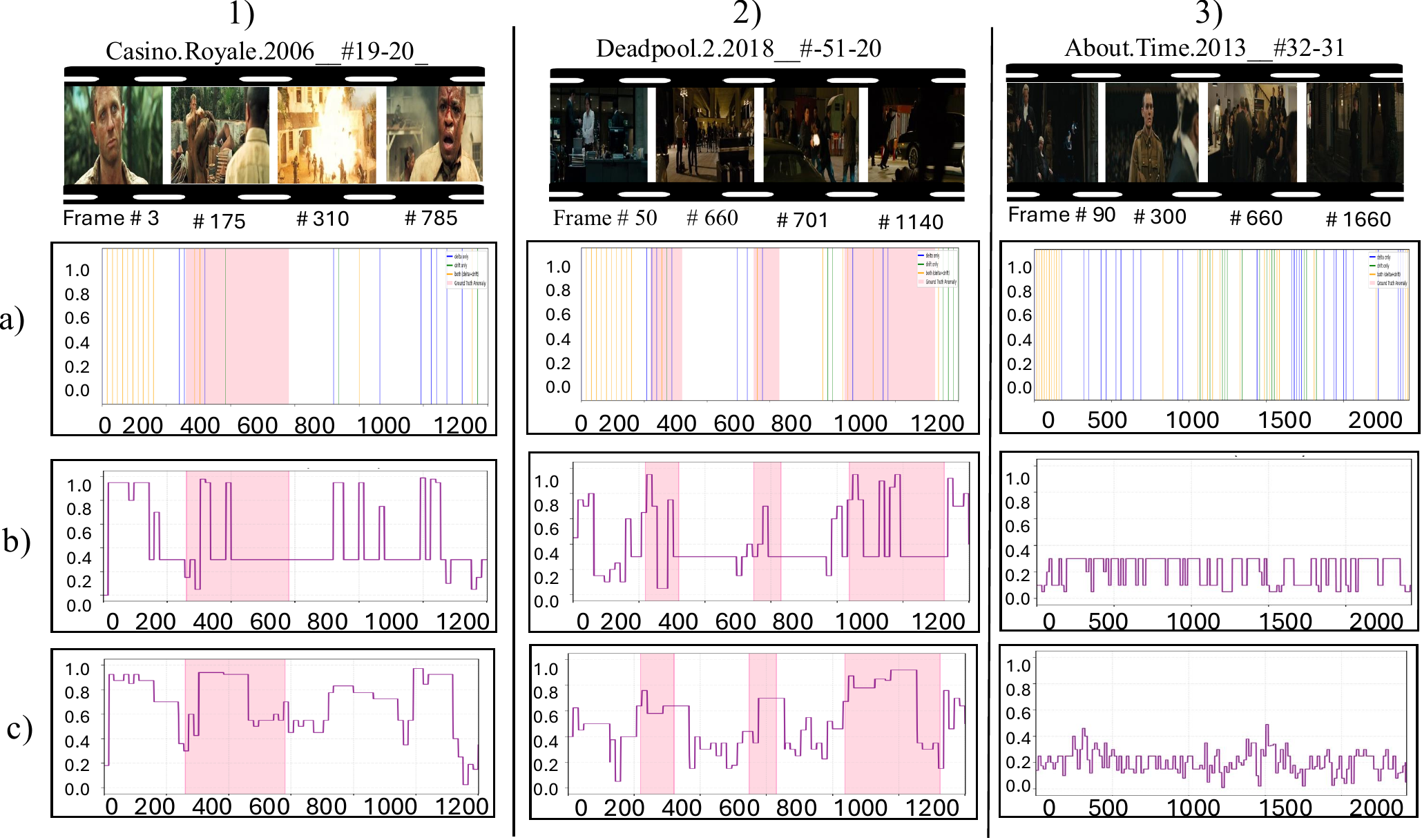}
    \caption{Shows Qualitative results. Each column shows: (a) signal activation patterns with ground-truth anomaly regions shaded in pink, blue vertical lines indicate delta-only activations, green lines indicate drift-only activations, and orange lines indicate frames where both signals are activated; (b) frame-level anomaly scores; and (c) video-level scores after threshold-gated fusion.}
  \label{fig:Qualitative}
  \vspace{-.5cm}
\end{figure}

\subsection{Qualitative results}
Figure \ref{fig:Qualitative} \textbf{Video 1:} an explosion spans frames 200--600. Signals (a) concentrate selection within the anomalous region, frame-level scores (b) rise sharply at onset and stay elevated, and gated fusion (c) yields a temporally coherent detection.
\textbf{Video 2:} three separate violent events. SAAF samples around each and skips the gaps (a); frame-level scores (b) peak at all three locations and fusion (c) sharpens the contrast against normal segments.
\textbf{Video 3:} a normal video with continuous motion drives persistent drift activation (a), yet SAAF selects only 43\% of frames. Both scoring stages (b, c) remain low throughout, avoiding false positives under high scene dynamics --
detailed success and failure cases are provided in (Supplementary~Sec.~B).

\begin{figure}[!b]
    \centering
    \includegraphics[width=\linewidth]{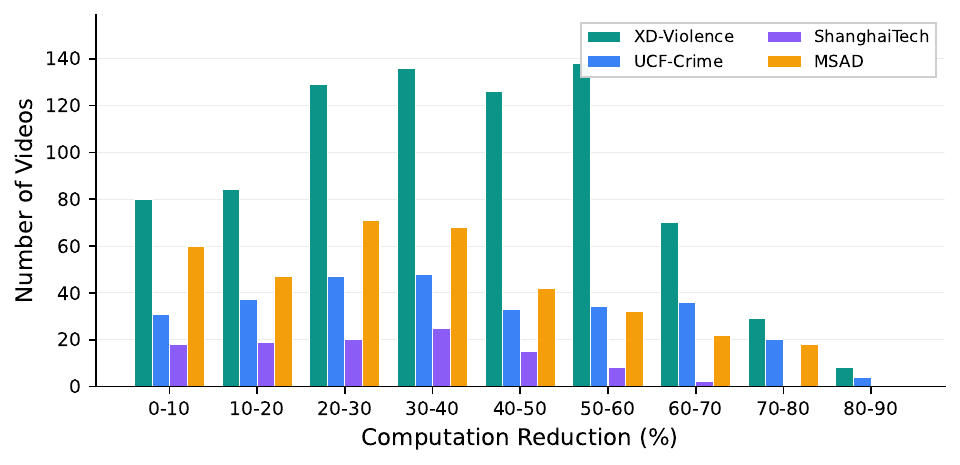}
    \caption{%
     Distribution of per-video VLM call reduction relative to uniform stride-16 sampling across benchmark datasets. Higher reduction indicates fewer VLM calls within the stride-16 candidate pool. Adaptive frame selection achieves 20–60\% reduction for the majority of videos while preserving detection accuracy.}
    \label{fig:reduction_dist}
\end{figure}

\section{Computation Reduction Analysis}
Figure~\ref{fig:reduction_dist} shows the per-video distribution of VLM inference calls saved relative to uniform stride-16 sampling across the four benchmarks.
Two observations stand out. First, adaptive selection achieves non-trivial reductions across the majority of videos, with the bulk concentrated in the 20--60\% range. XD peaks at 50--60\% (138 videos, the modal bin), while UCF and MSAD spread more uniformly across 20--50\%. Second, the long tail extending to 80--90\% reduction shows that videos with predominantly static content yield substantial savings without sacrificing detection capability.
 
Table~\ref{tab:computational} summarizes the per-component timing breakdown measured on a single NVIDIA A100 40GB GPU.

\begin{table}[!t]
\centering
\caption{Per-component computational cost across benchmarks.}
\label{tab:computational}
\small
\setlength{\tabcolsep}{3pt}
\begin{tabularx}{\linewidth}{>{\footnotesize}Xcc cccc}
\toprule
\multirow{2}{*}{\footnotesize\textbf{Component}} &
\multirow{2}{*}{\textbf{B}} &
\multirow{2}{*}{\textbf{T(s)}} &
\multicolumn{4}{c}{\textbf{\# Frames processed}} \\
\cmidrule(lr){4-7}
 & & & \textbf{XD} & \textbf{UCF} & \textbf{MSAD} & \textbf{Shang} \\
\midrule
\multicolumn{7}{l}{\textit{Stage 1: Signal Extraction (candidate frames)}} \\
CLIP ViT-B/32      & 24 & 0.14 & 146{,}449 & 69{,}634 & 14{,}360 & 2{,}550 \\
\midrule
\multicolumn{7}{l}{\textit{Stage 2--3: Frame-Level Scoring (selected frames)}} \\
VLM (Qwen3-VL-8B)  & 24 & 8.19 & 102{,}066 & 53{,}283 & 9{,}353 & 1{,}898 \\
LLM (Qwen2.5-7B)   & 24 & 0.64 & 102{,}066 & 53{,}283 & 9{,}353 & 1{,}898 \\
\midrule
\multicolumn{7}{l}{\textit{Stage 4: Video-Level Fusion (gated anchors, $s^{\text{frame}}>\tau_f$)}} \\
VideoLLaMA3-7B     & 1  & 0.84 & 15{,}968  & 3{,}300  & 2{,}327 & 487 \\
LLM (Qwen2.5-7B)   & 24 & 0.64 & 15{,}968  & 3{,}300  & 2{,}327 & 487 \\
\midrule
\textbf{Total (h)}        & & & 14.4   & 6.35   & 1.54   & 0.33 \\
\textbf{Amortized FPS}    & & & 44.95  & 48.61  & 41.06  & 33.55 \\
\bottomrule
\end{tabularx}
\vspace{-0.5cm}
\end{table}
The pipeline operates at stride-16, with CLIP extraction (0.14s/batch of 24).
Dual-signal gating selects 102{,}066, 53{,}283, and 9{,}353 frames
(69.7\%, 76.5\%, 65.1\%) on XD, UCF, and MSAD for captioning
(Qwen3-VL-8B, 8.19s/batch) and LLM scoring (Qwen2.5-7B, 0.64s/batch).
At the video level, 15{,}968, 3{,}300, and 2{,}327 anchors pass the
$\tau_f$ gate and undergo fusion via VideoLLaMA3-7B (0.84s) followed by
LLM scoring (0.64s/batch).  Table~\ref{tab:throughput} compares SAAF against
Unified \cite{lin2025unified}. By amortizing inference across total frames, SAAF reaches 44.95, 48.61, and 41.06 FPS on
XD, UCF, and MSAD, delivers 2.43$\times$, 2.63$\times$, and
2.22$\times$ speedups over Unified 18.5 FPS.
 
\begin{table}[!t]
\centering
\caption{Throughput and accuracy across XD, UCF, and MSAD. Measured under identical settings.\footnotesize $^{\dagger}$LAVAD does not report MSAD.}
\label{tab:throughput}
\small
\setlength{\tabcolsep}{3pt}
\begin{tabular}{lcccccc}
\toprule
& \multicolumn{2}{c}{\textbf{XD}} & \multicolumn{2}{c}{\textbf{UCF}} & \multicolumn{2}{c}{\textbf{MSAD}} \\
\cmidrule(lr){2-3} \cmidrule(lr){4-5} \cmidrule(lr){6-7}
\textbf{Method} & FPS & AUC & FPS & AUC & FPS & AUC \\
\midrule
LAVAD $^{\dagger}$~\cite{zanella2024harnessing} & 13.0 & 85.36 & 13.0 & 80.28 &-- & -- \\
Unified~\cite{lin2025unified}      & 18.5 & 91.34 & 18.5 & \textbf{84.28} & 18.5 & 85.90 \\
\textbf{SAAF (Ours)}               & \textbf{44.95} & \textbf{92.07} & \textbf{48.61} & 82.59 & \textbf{41.06} & \textbf{86.70} \\
\bottomrule
\end{tabular}
\end{table}

\section{Ablation Study}

\begin{figure}[h]
\centering
\includegraphics[width=\columnwidth]{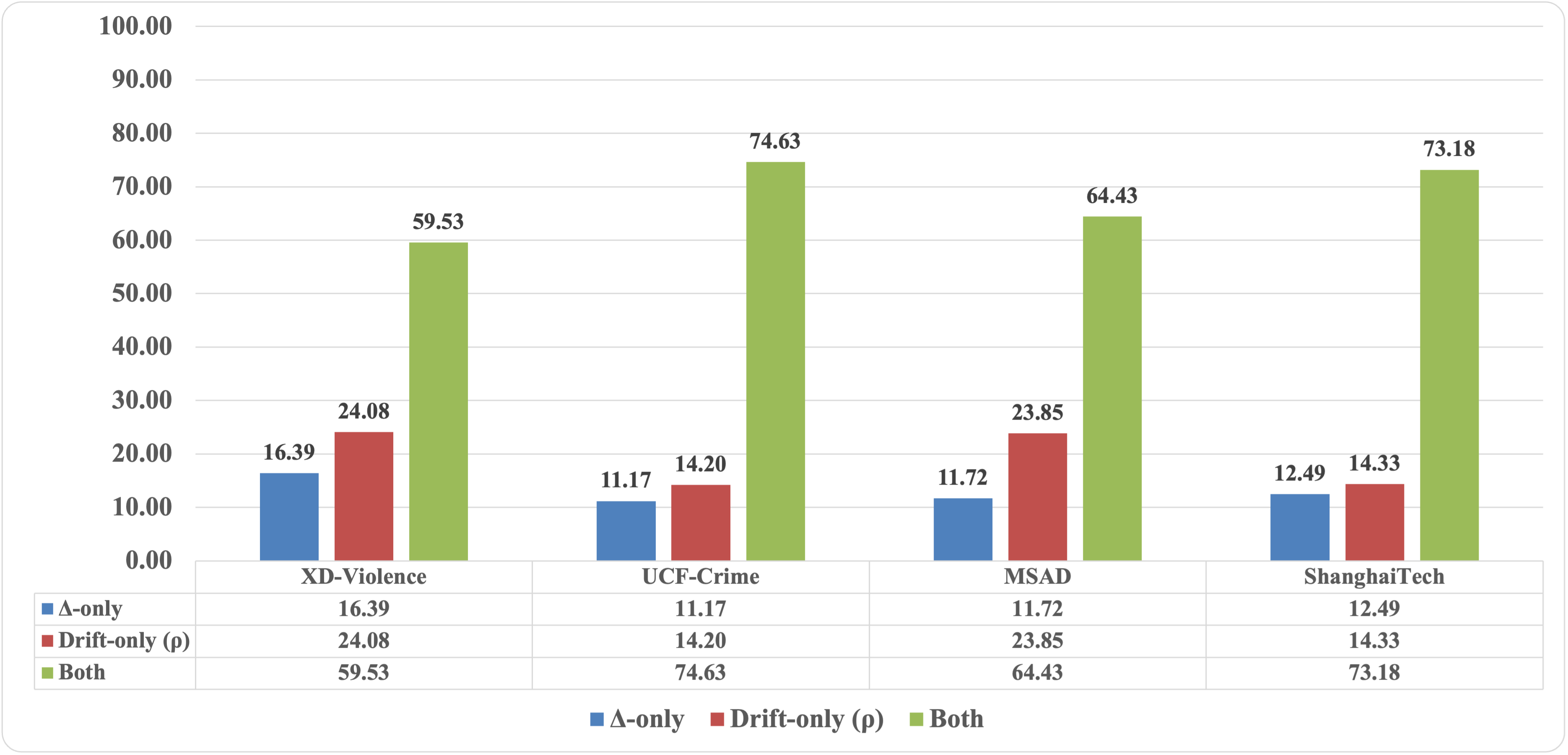}
\caption{Signal distribution (\%) among selected stride-16 frames.}
\label{fig:signal_attribution}
\vspace{-0.3cm}
\end{figure}

\noindent\textbf{Signal Attribution Distribution}
Figure~\ref{fig:signal_attribution} attributes each selected frame to $\delta$-only, $\rho$-only, or both. Across all four benchmarks, $\delta$-only accounts for 11--16\% of selections and $\rho$-only for 14--24\%, while 59--75\% trigger both signals simultaneously. The non-trivial single-signal proportions (25--40\% combined)
represent frames a single-signal approach would miss, directly explaining the performance margin in Supplementary Sec.~A.1 (Table~\ref{tab:signal_ablation}).

\noindent\textbf{Backbone Selection for Signal Extraction}
We evaluate four pretrained visual encoders for computing delta and drift signals. 
Table~\ref{tab:backbone_ablation} reports AUC-ROC under identical pipeline settings.
\begin{table}[t]
\centering
\caption{Backbone ablation for signal extraction on XD. All encoders achieve comparable performance. Drift window $W = 20$}
\label{tab:backbone_ablation}
\begin{tabular}{lc}
\toprule
Backbone & AUC-ROC (\%) \\
\midrule
CLIP ViT-B/32 & \textbf{92.07} \\
DINOv3 ViT-B/14 & 91.65 \\
DINOv2 ViT-B/14 & 91.78 \\
SigLIP2 ViT-B/16 & 91.61 \\
\bottomrule
\end{tabular}
\vspace{-0.3cm}
\end{table}
All backbones achieve comparable performance, with AUC-ROC spanning a narrow 0.46\% range. CLIP marginally outperforms others, while Dinov2 ViT-B/14 trails by only 0.29\%. This shows that the dual-signal mechanism generalizes across embedding spaces, capturing anomaly-relevant temporal patterns regardless of the underlying encoder. We adopt CLIP ViT-B/32 as the default.

\begin{table}[t]
\centering
\caption{Ablation on rolling window size $W$ for drift signal.}
\label{tab:window_ablation}
\begin{tabular}{cc}
\toprule
Window Size ($W$) & AUC-ROC (\%) \\
\midrule
5 & 91.77 \\
10 & 91.79 \\
15 & 91.80 \\
20 & \textbf{92.07} \\
\bottomrule
\end{tabular}
\vspace{-0.3cm}
\end{table}

\noindent\textbf{Rolling Window Size for Drift Signal}
The drift signal computes deviation from a rolling temporal anchor spanning $W$ preceding frames. We evaluate window sizes $W \in \{5, 10, 15, 20\}$ all others fixed. Table~\ref{tab:window_ablation} reports the results.
Performance improves slightly as window size increases. Smaller windows ($W = 5$) respond quickly to scene changes but produce noisier drift estimates, larger windows ($W \geq 20$) accumulate sufficient temporal context to distinguish genuine drift from short-term fluctuations.  We set $W = 20$ as the default.

\noindent\textbf{Stage-Wise Breakdown Analysis}
Table~\ref{tab:stage_ablation} reports cumulative AUC-ROC across pipeline stages. Each stage yields monotonic gains. $\lambda$-weighted video-level gating contributes +10.23\% on XD and +11.57\% on UCF, and refinement adds a further +6.32\% on UCF and +5.04\% on Shang.
\begin{table}[h]
\centering
\small
\caption{Stage-wise ablation across benchmarks. AUC-ROC (\%).}
\label{tab:stage_ablation}
\vspace{-0.3cm}
\resizebox{\columnwidth}{!}{
\begin{tabular}{lcccc}
\toprule
\textbf{Stage} & \textbf{XD} & \textbf{UCF} & \textbf{MSAD} & \textbf{Shang} \\
\midrule
2. Frame-level & 78.30 & 64.70 & 72.13 & 71.38 \\
3. + $\lambda$-gating & 88.53 & 76.27 & 82.36 & 75.50 \\
4. + Refinement & \textbf{92.07} & \textbf{82.59} & \textbf{86.70} & \textbf{80.54} \\
\bottomrule
\end{tabular}
}
\vspace{-0.3cm}
\end{table}

\noindent\textbf{Ablation of the Refinement Stage}
\label{sec:abl_refinement}
To isolate the contribution of conditional video-level fusion from Gaussian smoothing, we apply smoothing directly to Stage~2 outputs. Smoothing alone improves AUC-ROC on every benchmark, yet the full Stage~3 pipeline retains margins of $2.72$, $9.29$, $4.92$, and $3.35$ points on XD, UCF, MSAD, and Shang. The gap shows that video-level fusion supplies semantic evidence beyond what smoothing can provide.
\begin{table}[h]
\centering
\small
\setlength{\tabcolsep}{4pt}
\caption{Ablation of refinement components.}
\label{tab:abl_refinement}
\scalebox{0.93}{
\begin{tabular*}{\columnwidth}{@{\extracolsep{\fill}}lcccc@{}}
\toprule
Configuration & XD-Viol. & UCF & MSAD & Shang \\
\midrule
Stage~2 (no smoothing)        & 78.30 & 64.70 & 72.13 & 71.38 \\
Stage~2 + smoothing           & 89.35 & 73.30 & 81.78 & 77.19 \\
Stage~3 + smoothing (Stage~3) & \textbf{92.07} & \textbf{82.59} & \textbf{86.70} & \textbf{80.54} \\
\bottomrule
\end{tabular*}}
\vspace{-0.5cm}
\end{table}

\section{Conclusion}
We presented SAAF, an efficient zero-shot VAD framework that resolves the tension between semantic capability and computational feasibility. Among training-free zero-shot methods, SAAF achieves SOTA on XD, MSAD, and Shang while reducing computation, suggesting that content-aware allocation guided by lightweight semantic signals is a promising direction for deploying VLMs at scale.

{\small
\bibliographystyle{ieeenat_fullname}
\bibliography{main}
}

\clearpage
\appendix
\renewcommand{\thesection}{\Alph{section}}

\twocolumn[
\begin{center}
\Large\textbf{Supplementary Material for\\[0.5em] Efficient Zero-Shot Video Anomaly Detection via Signal-Aware Adaptive Fusion (SAAF)}
\end{center}
\vspace{1em}
]

\section*{Supplementary Roadmap}

This supplementary document provides extended experimental analysis, qualitative results, and implementation details that complement the main paper. We organize the material as follows:

\begin{itemize}
    \item \textbf{Section A} Additional ablation study.
    
    \item \textbf{Section B} Additional qualitative results through representative success and failure cases.
    
    \item \textbf{Section C} Prompt templates used throughout the pipeline.
    
    \item \textbf{Section D} Limitations.
    
    \item \textbf{Section E} Broader impact considerations.
\end{itemize}

\section{Additional ablation study}
\label{sec:ablations}

We conduct systematic ablations on all four benchmarks to evaluate the contribution of each SAAF component.

\subsection{Event-Level Recall of Stage 1 Selection}
\label{sec:stage1_recall}

We measure how many true anomalies Stage 1 discards before scoring, independently
of AUC-ROC. Using the dual-signal ($\delta \vee \rho$ on CLIP, $N=1$,
$W=20$, stride 16) and per-video binary ground truth, we define an \emph{event}
as a contiguous run of anomalous frames. An event is \emph{hit} when at least one
selected frame, the only passed to the VLM, falls within it, and
\emph{missed} otherwise. We separate misses by cause. An event is \emph{unhittable} when the stride-16
grid places no candidate frame inside it, so the selector never sees it; this
affects any stride-16 method equally. A \emph{SAAF miss} is when a candidate
frame was available but neither $\delta$ nor $\rho$ fired, which is the only
case the selector is responsible for.

Table~\ref{tab:stage1_recall} 
report the results.
On MSAD, UCF, and XD, Stage 1 recovers 98.3\%, 92.3\%, and 88.9\%
of events, and nearly all misses are selector decisions rather than sampling
blind spots. ShanghaiTech is the exception at 77.8\%, but 22 of its 43 missed
events are unhittable under stride 16, reflecting the dataset's brief, gradually
evolving anomalies rather than a thresholding failure, consistent with the coarse
temporal resolution noted in the main paper.

\begin{table}[t]
\centering
\small
\caption{Event-level recall of Stage 1 selection. \emph{Unhittable} events
contain no stride-16 candidate; \emph{SAAF miss} events had a candidate but no
signal crossed threshold.}
\label{tab:stage1_recall}
\setlength{\tabcolsep}{4pt}
\begin{tabular}{lccccc}
\toprule
Dataset & Events & Recall (\%) & Miss & Unhittable & SAAF miss \\
\midrule
MSAD          & 240  & 98.3 & 4   & 1  & 3   \\
UCF     & 156  & 92.3 & 12  & 0  & 12  \\
XD   & 1210 & 88.9 & 134 & 0  & 134 \\
Shang  & 194  & 77.8 & 43  & 22 & 21  \\
\bottomrule
\end{tabular}
\end{table}

\subsection{Signal Component Ablation}

Table~\ref{tab:signal_ablation} isolates the contribution of each signal in the adaptive frame selection. We evaluate four configurations: $\delta$-only (frame-to-frame change), $\rho$-only (context deviation), OR-gated (either signal triggers selection), and AND-gated (both signals trigger).

\begin{table}[h]
\centering
\small
\caption{Signal component ablation (Stage~1 AUC-ROC \%). OR-gated consistently outperforms single-signal variants.}
\label{tab:signal_ablation}
\setlength{\tabcolsep}{3pt}
\renewcommand{\arraystretch}{0.95}

\begin{tabular*}{\linewidth}{@{\extracolsep{\fill}}lcccc@{}}
\toprule
\textbf{Dataset} & \textbf{Delta} & \textbf{Drift} & \textbf{OR} & \textbf{AND} \\
\midrule
XD-Violence & 72.89 & 74.87 & \textbf{78.30} & 68.64 \\
UCF-Crime & 63.60 & 63.83 & \textbf{64.70} & 62.62 \\
MSAD & 67.11 & 70.42 & \textbf{72.13} & 64.95 \\
ShanghaiTech & 70.12 & 69.60 & \textbf{71.38} & 68.02 \\
\bottomrule
\end{tabular*}
\end{table}

The results validate three design choices. First, OR-gated combination consistently achieves the highest performance across all datasets, confirming that delta and drift signals capture complementary anomaly patterns. Second, AND-gated combination yields the worst results, as requiring both signals to fire misses anomalies that manifest through only one signature. Third, the relative contribution of each signal varies by dataset: drift-only outperforms delta-only on XD and MSAD where gradual anomalies are prevalent, while performance is comparable on UCF and Shang. This variation reflects the differing characteristics of anomalies across benchmarks.

\begin{table}[b]
\centering
\caption{Selection threshold ablation. Stage~1 AUC-ROC (\%) with selection rate relative to uniform stride-16 sampling. Lower $N$ yields higher selection rates but diminished efficiency gains.}
\label{tab:selrate_auc}
\setlength{\tabcolsep}{8pt}
\renewcommand{\arraystretch}{1.15}
\begin{tabular}{lccc}
\toprule
\textbf{Dataset} & \textbf{$N$} & \textbf{Sel. Rate (\%)} & \textbf{AUC-ROC} \\
\midrule
\multirow{4}{*}{XD-Violence}
  & 0.5 & 79.80 & \textbf{80.74} \\
  & 1.0 & 69.70 & 78.30 \\
  & 1.5 & 58.50 & 74.66 \\
  & 2.0 & 42.70 & 68.60 \\
\midrule
\multirow{4}{*}{UCF-Crime}
  & 0.5 & 83.90 & \textbf{64.87} \\
  & 1.0 & 76.50 & 64.70 \\
  & 1.5 & 67.50 & 64.25 \\
  & 2.0 & 45.70 & 59.18 \\
\midrule
\multirow{4}{*}{MSAD}
  & 0.5 & 76.40 & \textbf{75.23} \\
  & 1.0 & 65.10 & 72.13 \\
  & 1.5 & 61.10 & 72.37 \\
  & 2.0 & 55.40 & 71.61 \\
\midrule
\multirow{4}{*}{ShanghaiTech}
  & 0.5 & 82.10 & \textbf{73.46} \\
  & 1.0 & 72.40 & 71.38 \\
  & 1.5 & 65.70 & 70.86 \\
  & 2.0 & 59.40 & 69.94 \\
\bottomrule
\end{tabular}
\end{table}

\subsection{Selection Threshold ($N$)}

The parameter $N$ in the adaptive thresholding formula $\tau_{x,t} = \mu_x^{(t)} + N \cdot \sigma_x^{(t)}$ controls detection sensitivity by determining how many standard deviations above the running mean a signal must exceed to trigger frame selection. Table~\ref{tab:selrate_auc} reports both selection rate (relative to uniform stride-16 sampling) and AUC-ROC across $N \in \{0.5, 1.0, 1.5, 2.0\}$.
Lower $N$ produces more permissive thresholds and higher selection rates, while higher $N$ restricts selection to the most extreme deviations. The results reveal a coverage accuracy trade-off inherent to adaptive sampling. At $N = 0.5$, permissive thresholds achieve the highest AUC-ROC across all datasets but select 76--84\% of candidate frames, reducing the computational savings that motivate adaptive selection. At $N = 2$, aggressive filtering reduces selection to 43--59\% but misses informative frames, with XD suffering the largest degradation relative to the default ($-9.70$ points, from 78.30 to 68.60). We adopt $N = 1$ as the default, achieving a practical balance: selection rates of 65--77\% preserve the majority of anomaly-relevant frames while still providing efficiency gains over uniform sampling.

\subsection{Default Fill Value ($s_d$)}

Frames that are not selected by the dual-signal mechanism receive a default score $s_d$ encoding a normality prior. Table~\ref{tab:fill_value} evaluates the sensitivity of Stage~1 performance to $s_d \in \{0, 0.1, 0.2, 0.3\}$.

\begin{table}[h]
\centering
\caption{Default fill value ($s_d$) ablation. Stage~1 AUC-ROC (\%).}
\label{tab:fill_value}
\begin{tabular*}{\columnwidth}{@{\extracolsep{\fill}}ccccc@{}}
\toprule
$s_d$ & XD-Viol. & UCF & MSAD & Shang \\
\midrule
0 & 64.90 & 55.44 & 67.50 & 63.35 \\
0.1 & 76.26 & 61.52 & \textbf{72.74} & 68.15 \\
0.2 & \textbf{78.98} & \textbf{64.90} & 72.13 & \textbf{71.38} \\
0.3 & 78.30 & 64.70 & 72.13 & \textbf{71.38} \\
\bottomrule
\end{tabular*}
\end{table}

Setting $s_d = 0$ yields substantially degraded performance across all datasets, as unselected normal frames receive scores identical to confident normality predictions, compressing the score distribution and harming discriminability. Performance improves markedly with $s_d \geq 0.1$, which provides separation between unselected frames and explicitly scored normal content. The optimal value varies slightly by dataset: $s_d = 0.2$ achieves the best results on XD-Violence, UCF-Crime, and ShanghaiTech, while MSAD peaks at $s_d = 0.1$. We adopt $s_d = 0.3$ as the default, which performs within 0.7\% of optimal across all benchmarks while encoding a conservative normality prior that avoids over-penalizing unselected frames.

\subsection{Fusion Threshold Sensitivity ($\tau_f$).}
We evaluate the effect of the fusion threshold $\tau_f$ on detection performance and computational cost across all four benchmarks. Table~\ref{tab:fusion_threshold} reports AUC-ROC and the video-level invocation rate, the fraction of frames triggering cascaded VLM inference, for $\tau_f \in \{0.4, 0.5, 0.6, 0.7, 0.8, 0.9\}$.
Performance degrades gracefully as $\tau_f$ increases. On XD-Violence, AUC-ROC decreases by only 0.86\% while the invocation rate drops from 10.9\% to 4.5\%, a $2.4\times$ reduction in VLM calls. UCF-Crime and MSAD remain within 0.7\%, and ShanghaiTech within 0.3\%, across the full sweep. The invocation rate on ShanghaiTech collapses at $\tau_f \geq 0.8$ (from 19.5\% to 0.1\%), as the video-level model assigns low confidence scores to its subtle anomalies, reducing SAAF to frame-level scoring.
We adopt $\tau_f = 0.4$ as the default to maximize recall, but $\tau_f$ can be raised for substantial efficiency gains at negligible accuracy cost, making SAAF adaptable to varying deployment budgets.

\begin{table}[h]
\centering
\caption{Effect of fusion threshold $\tau_f$ on AUC-ROC (\%) and invocation rate (\%). Higher $\tau_f$ reduces VLM calls with minimal performance degradation.}
\label{tab:fusion_threshold}
\resizebox{\columnwidth}{!}{%
\begin{tabular}{l cc cc cc cc}
\toprule
& \multicolumn{2}{c}{XD-Violence} & \multicolumn{2}{c}{UCF-Crime} & \multicolumn{2}{c}{MSAD} & \multicolumn{2}{c}{ShanghaiTech} \\
\cmidrule(lr){2-3} \cmidrule(lr){4-5} \cmidrule(lr){6-7} \cmidrule(lr){8-9}
$\tau_f$ & AUC & Inv.\% & AUC & Inv.\% & AUC & Inv.\% & AUC & Inv.\% \\
\midrule
0.4 & \textbf{88.53} & 10.90 & \textbf{76.27} & 4.7 & \textbf{82.36} & 16.2 & 75.50 & 22.5 \\
0.5 & 88.00 & 10.5 & 75.86 & 3.4 & 82.08 & 16.2 & 75.35 & 22.5 \\
0.6 & 88.00 & 10.5 & 75.86 & 3.4 & 82.07 & 16.1 & 75.35 & 22.5 \\
0.7 & 87.87 & 8.5 & 75.80 & 1.8 & 82.01 & 15.2 & \textbf{75.61} & 19.5 \\
0.8 & 87.81 & 7.6 & 75.85 & 1.5 & 81.88 & 12.1 & 75.47 & 0.1 \\
0.9 & 87.67 & 4.5 & 75.70 & 0.7 & 81.71 & 9.2 & 75.41 & 0.0 \\
\bottomrule
\end{tabular}%
}
\end{table}

\subsection{Signal-Aware Fusion Weights ($\lambda_\delta$, $\lambda_\rho$)}

The signal-aware fusion mechanism assigns weights $\lambda_\delta$ and $\lambda_\rho$ to frame-level scores based on which signal triggered selection: $\delta$-triggered frames weight frame-level scores by $\lambda_\delta$, while $\rho$-triggered frames use $\lambda_\rho$. When both signals fire, we apply $\lambda_{\text{both}} = 0.5$. Table~\ref{tab:fusion_weights} evaluates four weight configurations across all datasets.

\begin{table}[h]
\centering
\caption{Signal-aware fusion weight ablation. Stage~2 AUC-ROC (\%) under different $(\lambda_\delta, \lambda_\rho)$ configurations.}
\label{tab:fusion_weights}
\begin{tabular*}{\columnwidth}{@{\extracolsep{\fill}}lcccccc@{}}
\toprule
Config & $\lambda_\delta$ & $\lambda_\rho$ & XD & UCF & MSAD & Shang \\
\midrule
firm & 0.8 & 0.2 & 88.07 & 75.98 & 82.08 & 75.35 \\
moderate & 0.7 & 0.3 & 88.38 & 76.19 & 82.30 & 75.31 \\
default & 0.6 & 0.4 & \textbf{88.53} & \textbf{76.27} & \textbf{82.36} & 75.50 \\
reversed & 0.2 & 0.8 & 88.14 & 76.08 & 81.73 & \textbf{76.43} \\
\bottomrule
\end{tabular*}
\end{table}

We adopt $(\lambda_\delta, \lambda_\rho) = (0.6, 0.4)$ as our default. $\delta$ activations mark abrupt visual transitions whose evidence is concentrated in the triggering frame, so frame-level scoring dominates the fused decision. $\rho$ activations mark persistent or gradually evolving states whose interpretation benefits from temporal context, so the video-level score receives higher weight. This weighting aligns fusion with the semantics of the triggering signal.
Across XD-Violence, UCF-Crime, and MSAD, performance varies by less than 0.4 points across configurations, indicating the stage is robust to this choice. ShanghaiTech is the exception, where a $\rho$-dominant weighting (0.2, 0.8) yields a 0.93-point gain, reflecting the dataset's bias toward gradually evolving anomalies in a single scene. We retain (0.6, 0.4) as the default to avoid dataset-specific tuning.

\subsection{Additional Per-Category Analysis}
\begin{figure*}[t]
    \centering
    \includegraphics[width=\textwidth]{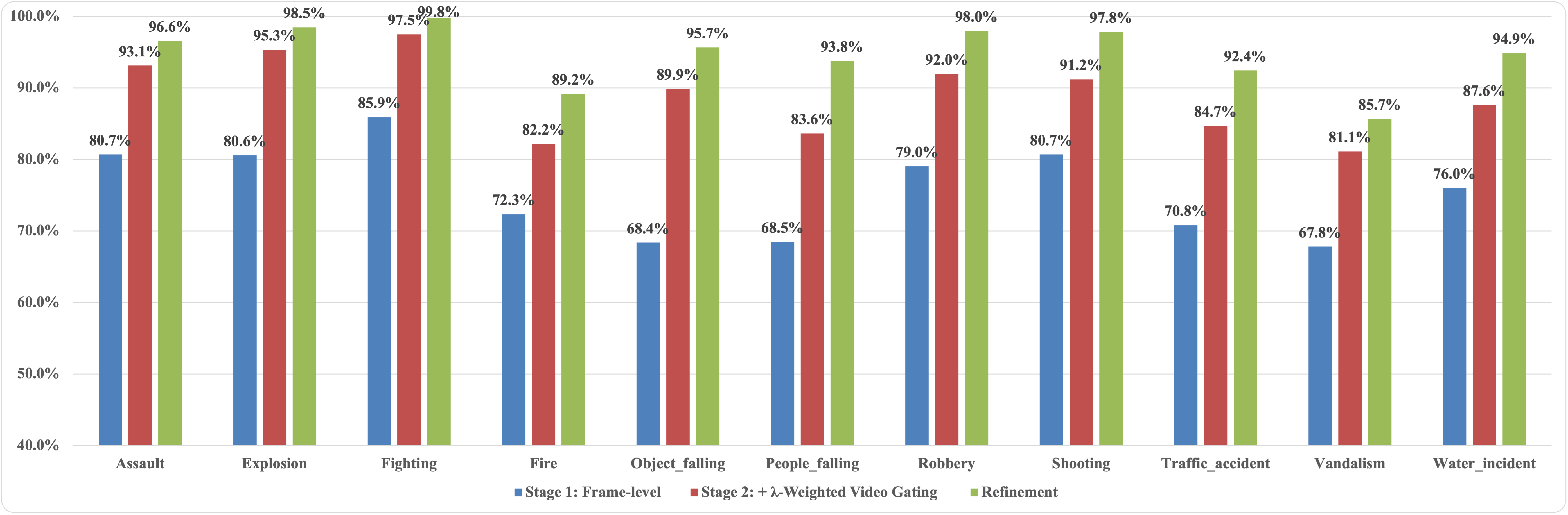}
    \caption{MSAD (11 anomaly categories)}
    \label{fig:breakdown-msad}
\end{figure*}
Figure~\ref{fig:breakdown-msad} presents the per-category AUC-ROC breakdown across the three SAAF pipeline stages on MSAD.

\textbf{Strong performance categories.} SAAF achieves near-perfect detection ($>$95\% AUC-ROC after refinement) on high-motion sudden events such as Fighting (99.8\%), Explosion (98.5\%), Robbery (98.0\%), Shooting (97.8\%), Assault (96.6\%), and Object\_falling (95.7\%). These categories exhibit strong visual signatures such as rapid motion, dramatic scene changes, or impacts that both the $\delta$ signal and frame-level VLM scoring capture reliably.

\textbf{Challenging categories.} Gradual-onset and visually subtle anomalies remain more difficult. Vandalism (85.7\%) involves fine-grained actions that may not produce distinctive $\delta$ or $\rho$ signals until the act is underway. Fire (89.2\%) is limited by slow visual progression that the drift signal must accumulate over time.

\textbf{Stage-wise improvement patterns.} The multi-stage architecture provides consistent gains across all categories. Video-level gating (Stage~2) delivers the largest improvements on categories requiring temporal context, with Object\_falling improving by 21.5\% and People\_falling by 15.1\%. Temporal refinement (Stage~3) further boosts performance by 2--10\% across categories. These patterns indicate that sudden events benefit from accurate frame-level detection, while ambiguous or gradual anomalies prefer video-level reasoning.

\begin{figure*}[t!]
    \centering
    
    \begin{subfigure}{0.48\textwidth}
        \centering
        \includegraphics[width=\linewidth]{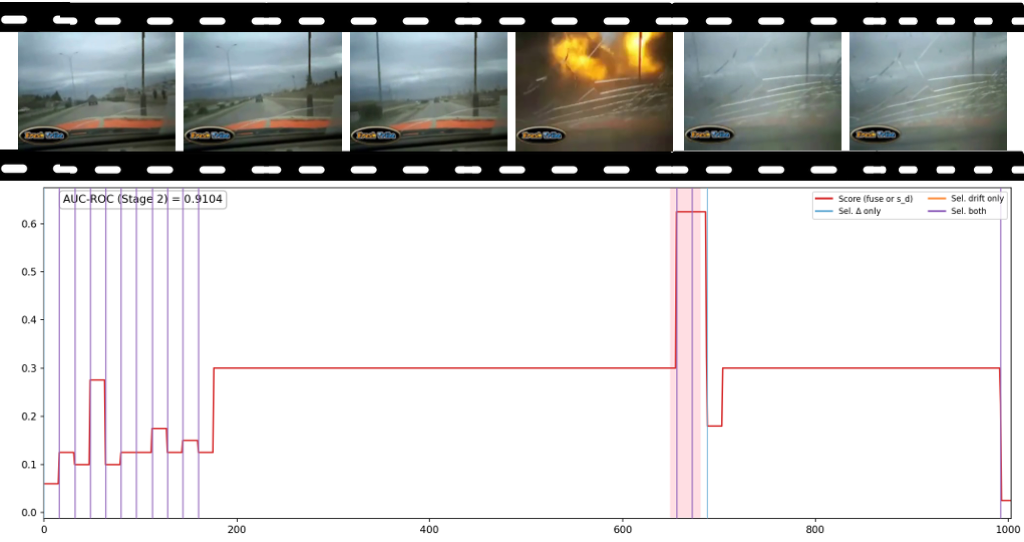}
        \caption{Success case: Road accident captured from dashcam (XD-Violence). SAAF concentrates frame selection around the anomalous event, achieving 76.2\% computational savings with high detection accuracy.}
    \end{subfigure}
    \hfill
    \begin{subfigure}{0.48\textwidth}
        \centering
        \includegraphics[width=\linewidth]{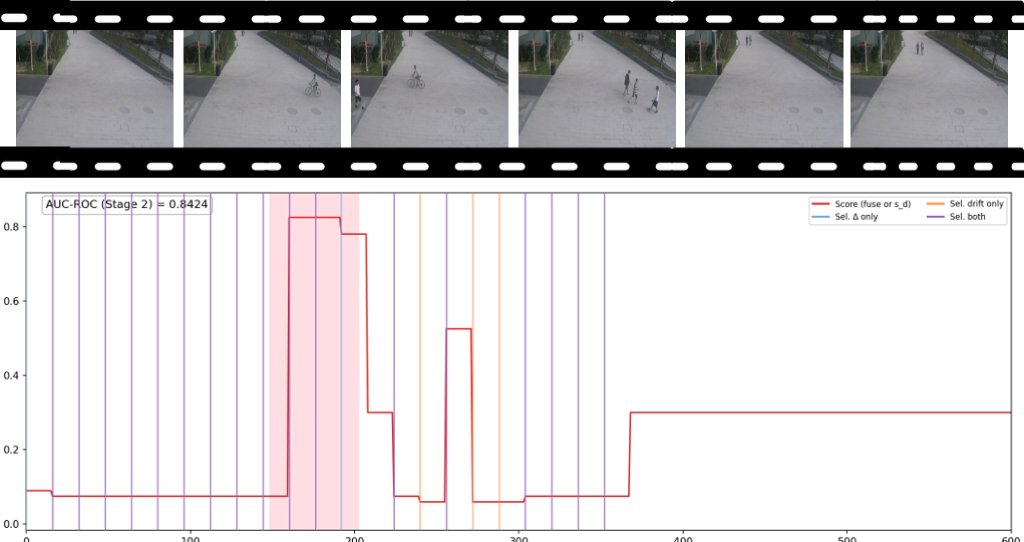}
        \caption{Success case: Cycling anomaly in pedestrian area (ShanghaiTech). SAAF accurately detects the anomalous activity while skipping normal pedestrian scenes, achieving 42.1\% computational savings.}
    \end{subfigure}
    
    \vspace{0.5cm}
    
    \begin{subfigure}{0.48\textwidth}
        \centering
        \includegraphics[width=\linewidth]{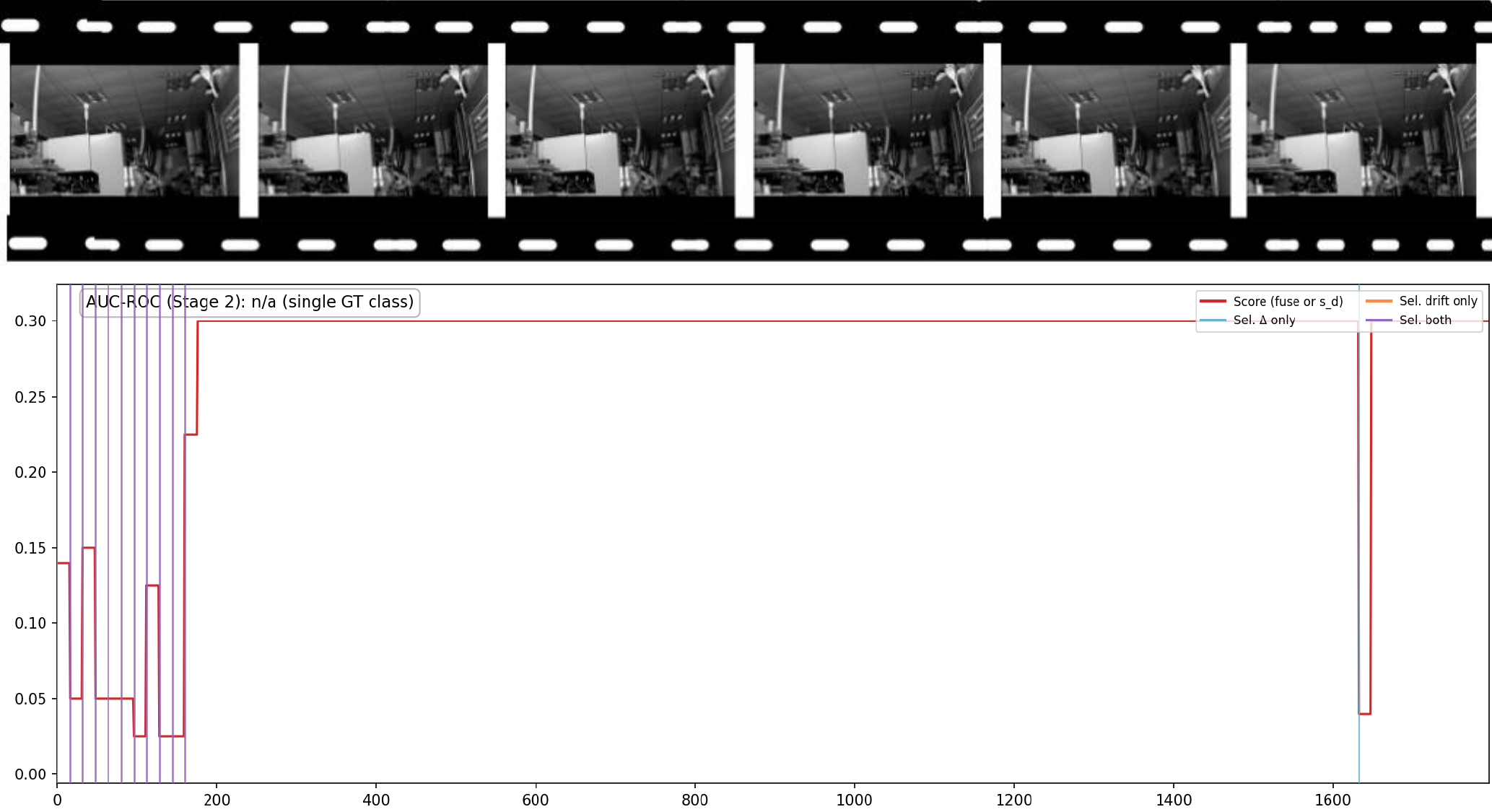}
        \caption{Success case: Normal video (UCF-Crime). SAAF correctly identifies this as a non-anomalous scene, maintaining low scores while achieving 89.4\% computational savings through smart frame selection.}
    \end{subfigure}
    \hfill
    \begin{subfigure}{0.48\textwidth}
        \centering
        \includegraphics[width=\linewidth]{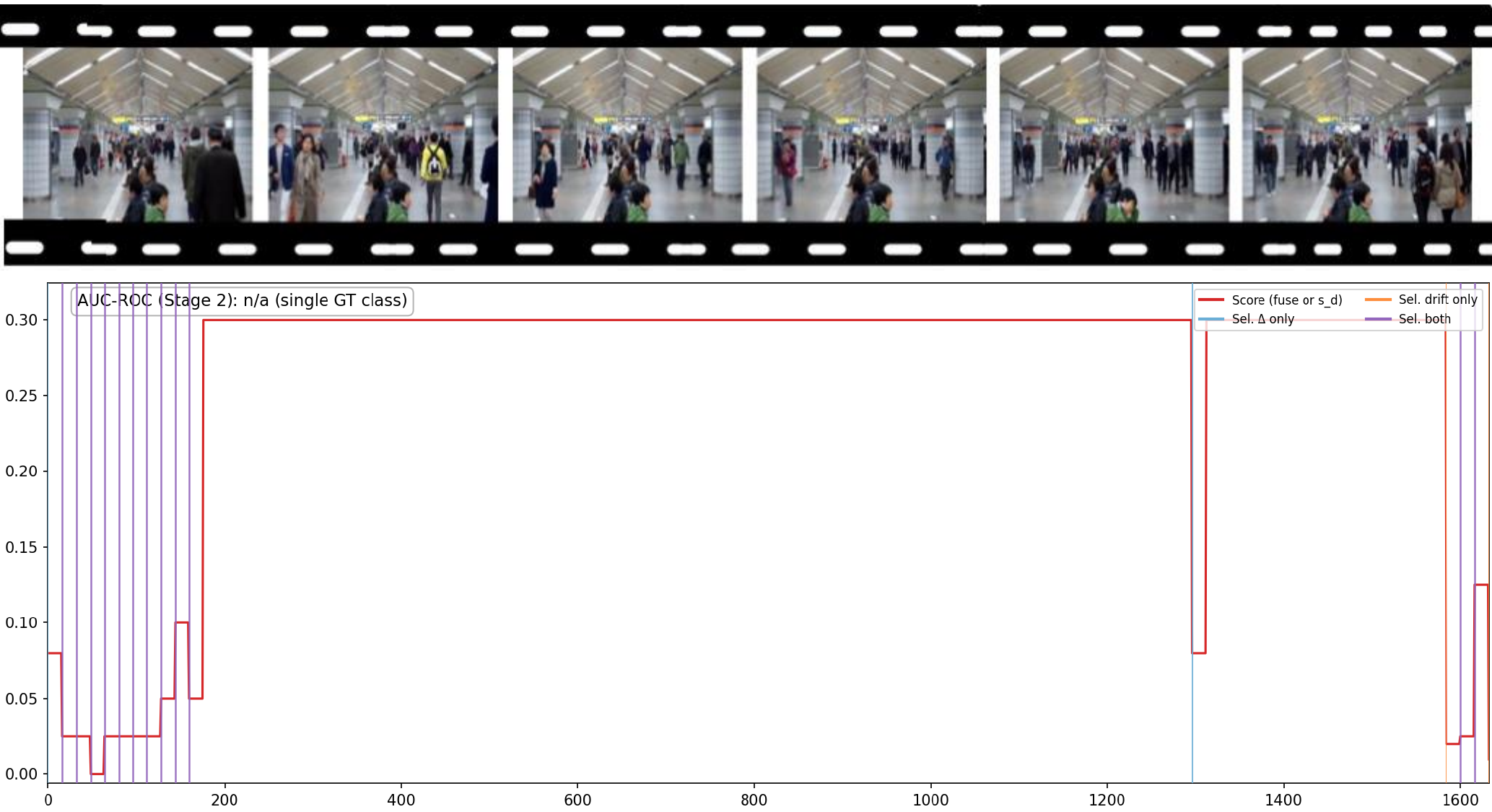}
        \caption{Success case: Normal video from metro station (XD-Violence). Despite continuous pedestrian activity, SAAF maintains low anomaly scores and achieves 84.5\% computational savings.}
    \end{subfigure}
    
    \caption{Representative success cases demonstrating SAAF's detection accuracy and computational efficiency across anomalous (a, b) and normal (c, d) videos.}
    \label{fig:four_subfigures_success}
\end{figure*}

\begin{figure*}[t]
    \centering
    
    \begin{subfigure}{0.48\textwidth}
        \centering
        \includegraphics[width=\linewidth]{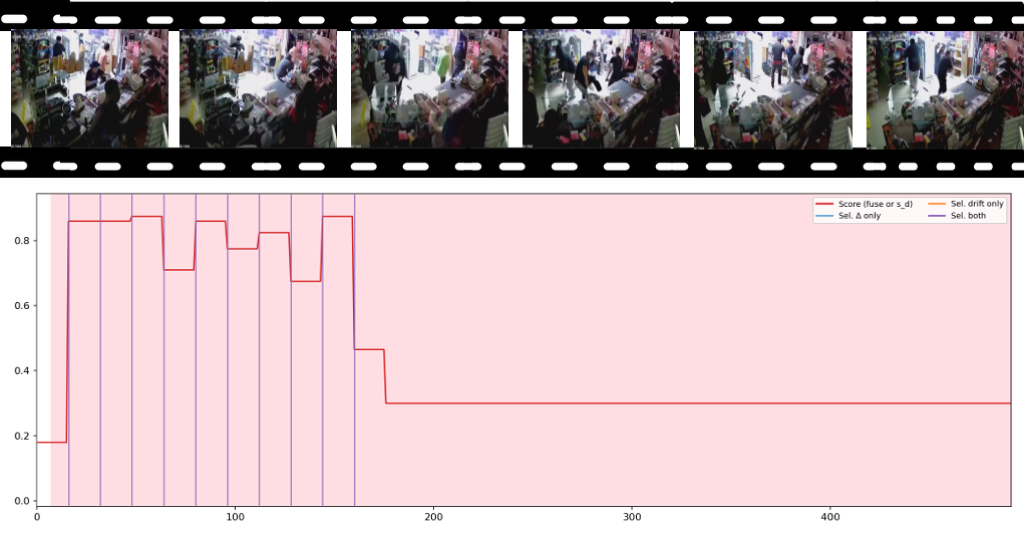}
        \caption{Failure case: Robbery\_1 (MSAD). The anomaly begins at frame 0, contaminating baseline statistics during warmup and preventing proper signal activation.}
    \end{subfigure}
    \hfill
    \begin{subfigure}{0.48\textwidth}
        \centering
        \includegraphics[width=\linewidth]{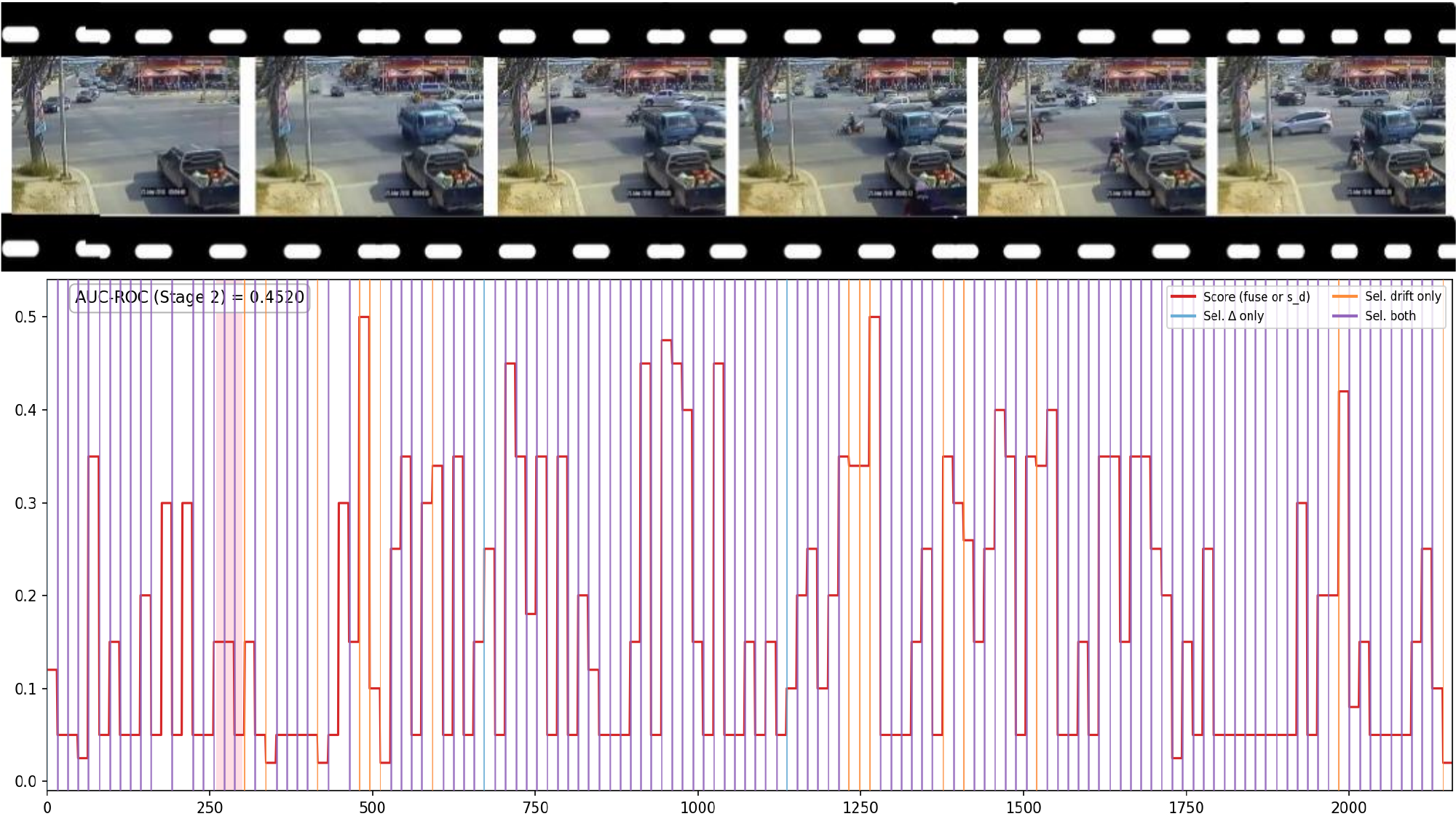}
        \caption{Failure case: RoadAccidents011 (UCF-Crime). Intense, fast-moving traffic triggers dense signal activations throughout the video. The limited computation savings are attributed to the fast-changing scene.}
    \end{subfigure}
    
    \vspace{0.5cm}
    
    \begin{subfigure}{0.48\textwidth}
        \centering
        \includegraphics[width=\linewidth]{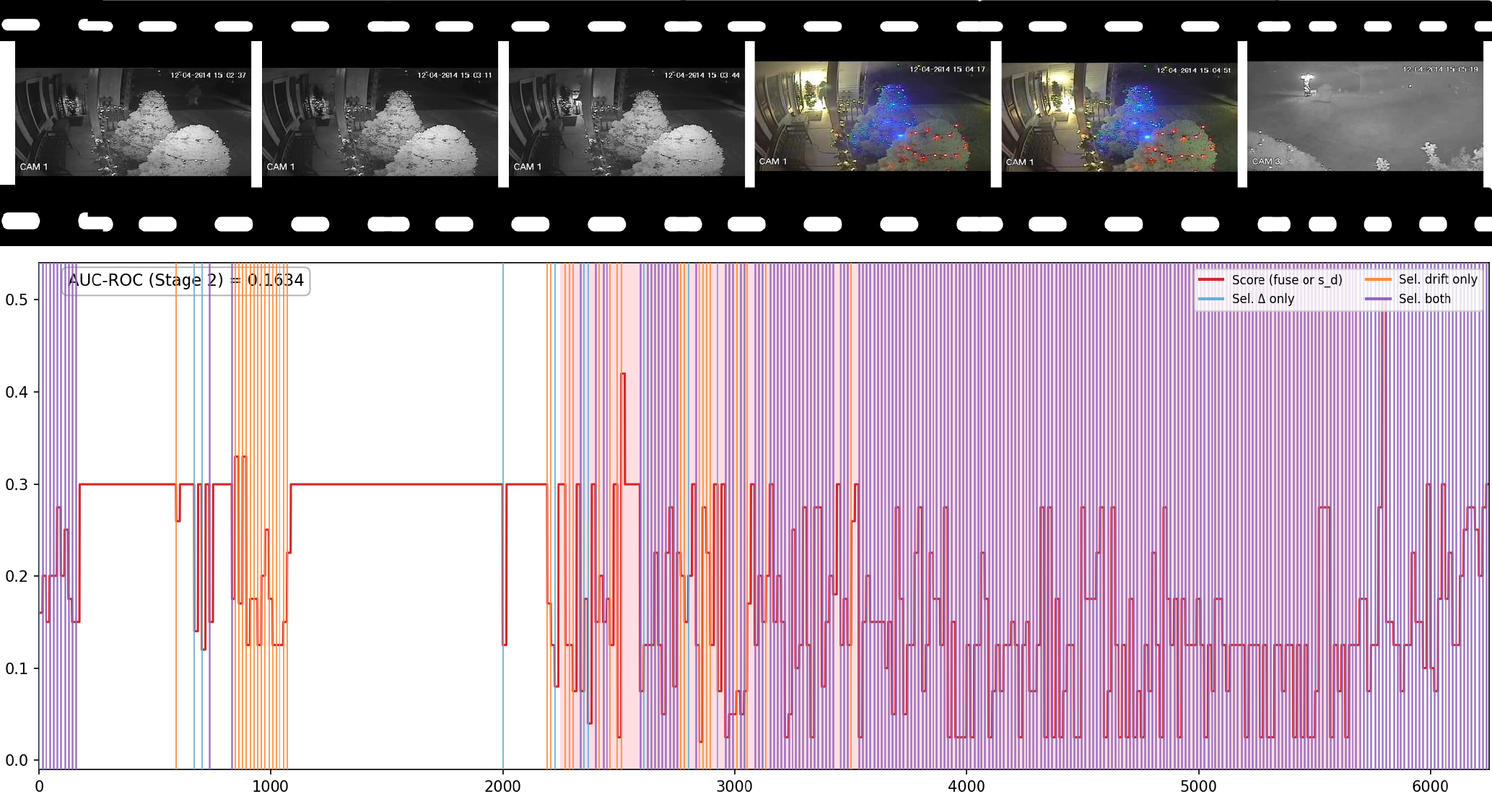}
        \caption{Failure case: Arson007 (UCF-Crime). The dual-signal mechanism correctly identifies the anomalous region, but the VLM fails to recognize the subtle fire progression.}
    \end{subfigure}
    \hfill
    \begin{subfigure}{0.48\textwidth}
        \centering
        \includegraphics[width=\linewidth]{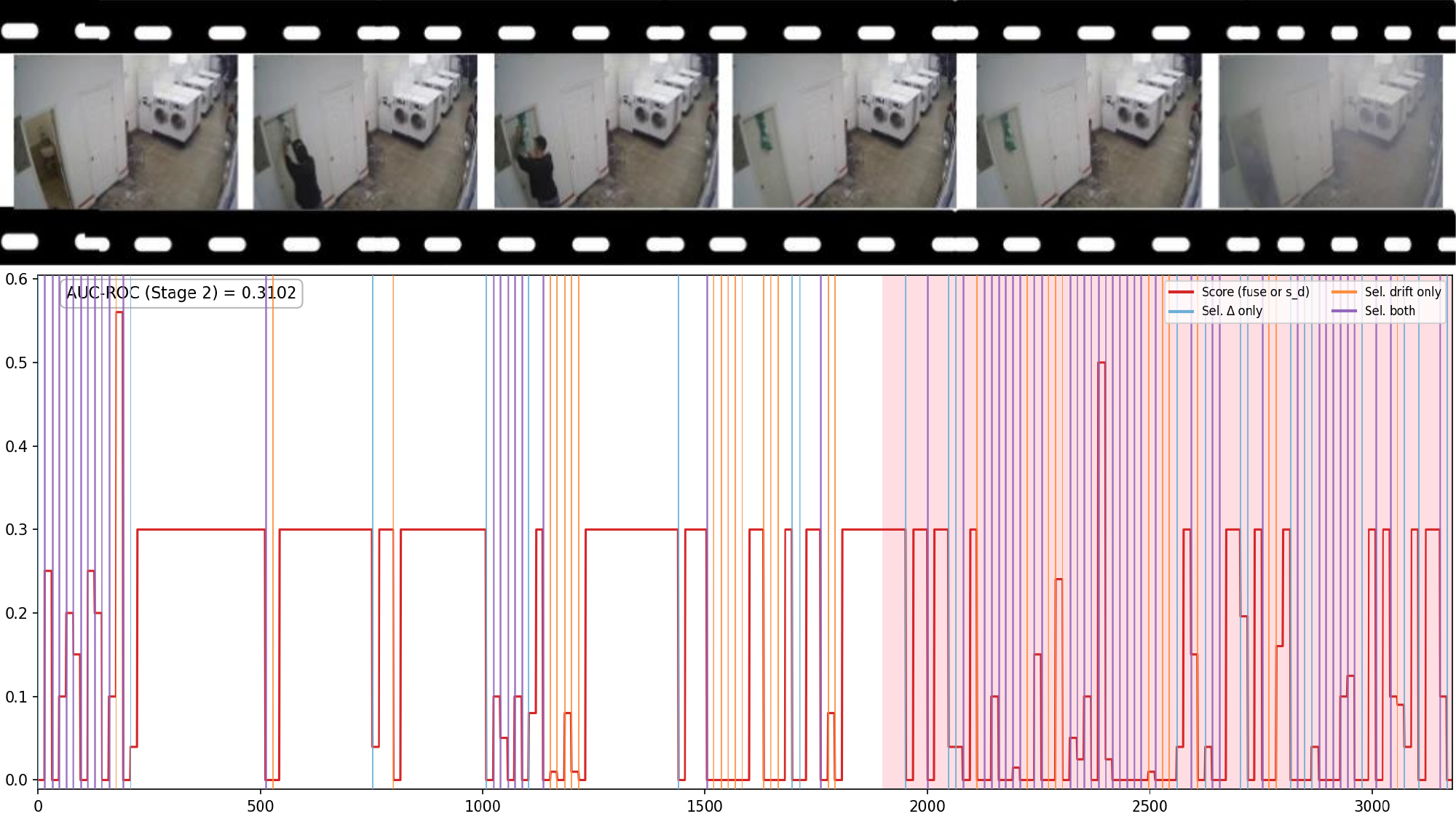}
        \caption{Failure case: Fire\_12 (MSAD). The dual-signal mechanism correctly concentrates selections in the anomalous region, but the VLM fails to distinguish subtle fire event.}
    \end{subfigure}
    
    \caption{Representative failure cases illustrating SAAF's limitations.}
    \label{fig:four_subfigures}
\end{figure*}

\section{Additional qualitative results}
This section presents qualitative analysis of SAAF's behavior through representative success and failure cases, illustrating both the framework's strengths and its current limitations.

\subsection{Qualitative Success Cases}

We present representative examples where SAAF successfully detects anomalies while achieving substantial computational savings. Figure~\ref{fig:four_subfigures_success} illustrates four spanning anomalous events and normal content across different datasets.

\textbf{Efficient detection of sudden events.} Figure~\ref{fig:four_subfigures_success}(a) demonstrates SAAF operating effectively on a dashcam-captured road accident from XD-Violence. The dual-signal mechanism correctly concentrates activations around the anomalous region (frames 650--750, shaded), where a sudden explosion and car accident occurs. The $\delta$ signal responds sharply to the  visual transition as the explosion ignites, while both signals appropriately skip the preceding normal driving footage (frames 200--630) where the scene remains static. The VLM produces a high anomaly score (0.65) precisely within the event region, providing clear separation from normal frames.

The computational efficiency is substantial, out of 63 candidate frames, only 15 are selected for stage1 and stage 2, a selection ratio of 23.8\%. This translates to 76.2\% computational savings compared to uniform stride-16 sampling, as the expensive VLM pipeline is invoked only when the dual-signal mechanism detects meaningful content. This case exemplifies SAAF's design goal, directing expensive computation toward anomaly-relevant content while aggressively skipping redundant frames, achieving both high detection accuracy and substantial efficiency gains.

\textbf{Detecting violations in pedestrian zones.} Figure~\ref{fig:four_subfigures_success}(b) shows a cycling anomaly from ShanghaiTech, where a bicycle enters a pedestrian-only area. The dual-signal mechanism responds robustly to the anomalous region (frames 160--220, shaded), with both $\delta$ and $\rho$ signals firing as the cyclist appears and moves through the scene. The Stage~2 scoring produces high anomaly scores (0.8--0.85) precisely within the event window, correctly identifying the violation. Notably, the signals also activate during the post-anomaly period (frames 250--350) as pedestrian activity increases, though scores appropriately decrease (0.3--0.55) reflecting absent anomaly presence. Out of 38 stride-16 candidate frames, 22 are selected for VLM analysis, a selection ratio of 57.9\%, yielding 42.1\% computational savings compared to uniform sampling.

\textbf{Efficient processing of normal content.} Figure~\ref{fig:four_subfigures_success}(c) demonstrates SAAF's behavior on a normal video from UCF-Crime containing no anomalous events (all-zero ground truth). The scene shows a static industrial environment with minimal activity across 1,600+ frames. The dual-signal mechanism correctly recognizes the lack of meaningful content: activations occur only during brief initial dynamics (frames 0--200) and minor scene changes near the end (frames 1,500--1,600), while the vast majority of frames are appropriately skipped.

The efficiency gains are substantial, only 12 out of 113 stride-16 candidate frames are selected for VLM analysis, a selection ratio of 10.6\%, yielding 89.4\% computational savings compared to uniform sampling. Throughout the video, anomaly scores remain consistently low, correctly reflecting the absence of anomalous content. 

\textbf{Handling dynamic normal scenes.} Figure~\ref{fig:four_subfigures_success}(d) presents a challenging normal video from XD-Violence depicting a busy metro station with continuous pedestrian flow. Unlike the static industrial scene in Figure~\ref{fig:four_subfigures_success}(c), this environment contains persistent motion from commuters walking, entering, and exiting the frame.

The dual-signal mechanism demonstrates appropriate calibration: activations concentrate during initial scene establishment (frames 0--200) and brief periods of elevated activity (frames 1,200--1,400), while the extended middle portion (frames 200--1,200) is correctly skipped despite ongoing pedestrian movement. Crucially, Stage 2 produces consistently low scores (0.02--0.08) for all selected frames, correctly classifying the normal commuter behavior.

The computational efficiency remains high even in this dynamic setting: only 16 out of 103 stride-16 candidate frames require Stage~1 \& Stage~2 analysis, a selection ratio of 15.5\%, yielding 84.5\% computational savings. This case demonstrates that SAAF's efficiency gains extend beyond static scenes to dynamic environments with regular activity patterns, invoking VLM inference only when scene dynamics deviate from established baselines rather than responding to all motion indiscriminately.

\subsection{Failure Case Analysis}

We analyze representative failure cases to understand SAAF's limitations and identify directions for future improvement. Figure~\ref{fig:four_subfigures} presents four scenarios that expose distinct failure modes.

\textbf{Baseline contamination from early-onset anomalies.} Figure~\ref{fig:four_subfigures}(a) reveals a structural limitation of SAAF's adaptive thresholding mechanism. The robbery event begins immediately at frame 0, meaning the anomalous content is present during the warmup period when baseline statistics ($\mu$ and $\sigma$) are initialized. As a result, the EMA-based thresholds adapt to the anomaly as if it were normal behavior, contaminating the baseline distribution. The consequence is visible in the signal activation pattern, despite the entire shaded region (frames 0--200) being anomalous, the dual-signal mechanism produces sparse activations because the $\delta$ and $\rho$ values do not exceed thresholds calibrated on anomalous content. Notably, it assigns high scores (0.7--0.9) for the frames that are selected, demonstrating that the scoring stage functions correctly, the failure lies entirely in frame selection due to corrupted baseline initialization. This case exposes an assumption of our approach, the warmup period must contain representative normal content to establish meaningful thresholds. Videos where anomalies begin immediately at the early frames violate this assumption, causing the adaptive mechanism to normalize abnormal patterns.

\textbf{Brief anomalies in high-motion environments.} Figure~\ref{fig:four_subfigures}(b) shows a road accident from UCF. The scene depicts a busy intersection with continuous vehicle and pedestrian movement, causing the dual-signal mechanism to activate densely throughout the entire video, both within and outside the brief anomalous region (frames 250--300, shaded). The accident itself, a brief collision lasting only ~50 frames, produces visual cues similar to vehicles braking or turning abruptly. This case illustrates SAAF's difficulty with transient anomalies embedded in dynamic scenes, where VLM ambiguity regarding subtle collision cues results in poor separation between anomalous and normal content.

\textbf{Subtle anomalies beyond VLM recognition.} Figure~\ref{fig:four_subfigures}(c) shows an arson case from UCF-Crime where $\delta$ and $\rho$ fire correctly within the anomalous region (frames 2,500--5,500, shaded) yet anomaly scores remain low throughout the event. The night-vision footage shows gradual fire progression with faint smoke and small flames against a cluttered background of Christmas lights, visual cues that fall below the VLM's detection capability. When anomalies exhibit signatures that challenge current VLMs, accurate selection alone cannot compensate.

\textbf{Visually subtle fire in industrial setting.} A similar pattern emerges in Figure~\ref{fig:four_subfigures}(d), which shows a fire event from MSAD occurring in an industrial environment with HVAC equipment. The signal activation pattern confirms that SAAF's frame selection operates correctly: the dual-signal mechanism appropriately concentrates activations within the anomalous region (shaded) where fire develops. However, the anomaly scores remain low throughout the event, failing to distinguish the fire from normal frames.

\section{Prompts and Templates}
\label{sec:prompts}

This section provides the prompt templates used throughout the SAAF pipeline. All prompts are designed to elicit structured, task-relevant outputs from the respective models while maintaining consistency across inference stages. We present three core prompts: the frame-level captioning prompt ($p_{\text{cap}}$) for visual description, the anomaly scoring prompt ($p_{\text{score}}$) for classification reasoning, and the video-level temporal prompt ($p_{\text{vid}}$) for multi-frame analysis. We conclude with examples demonstrating the inference chain.
\subsection{Frame-Level Visual Captioning Prompt}
The captioning prompt $p_{\text{cap}}$ instructs Qwen3-VL-8B to generate comprehensive scene descriptions with emphasis on relevant details. The prompt is structured to extract information across multiple dimensions relevant to anomaly detection.
\begin{tcolorbox}[
  colback=gray!3!white,
  colframe=gray!60!black,
  title=Frame-Level Captioning Prompt ($p_{\text{cap}}$),
  fonttitle=\bfseries
]
\small
\texttt{Analyze this footage frame and provide scene description:}\\[0.5em]
\texttt{\quad - All people present: their count, positions, activities, and interactions}\\
\texttt{\quad - Any unusual behaviors or events}\\
\texttt{\quad - Objects of interest: bags, weapons, vehicles, or suspicious items}\\
\texttt{\quad - Environmental context: location type (street, pedestrian area, parking lot, etc.}
\\
\texttt{\quad - Vehicles in pedestrian areas: note any cars, bicycles, skateboards, or carts in pedestrian-only zones}\\
\texttt{\quad - Any safety or security concerns}
\end{tcolorbox}
The structured format encourages the VLM to attend to specific visual elements that correlate with anomalous activity: person count and behavior, object presence, environmental context, and potential safety concerns. This decomposition ensures that downstream reasoning receives rich descriptions rather than generic scene summaries.
\subsection{Anomaly Scoring Prompt}

The scoring prompt $p_{\text{score}}$ provides Qwen2.5-7B-Instruct with explicit criteria for anomaly classification, used for both frame-level and video-level. The prompt enumerates concrete indicators for both anomalous and normal scenarios.
The enumeration of specific anomaly categories aligns with the types present in the evaluation benchmarks (XD, UCF, MSAD, ShanghaiTech), while the normal indicators help calibrate the model's baseline expectations. The LLM outputs a scalar score $s \in [0, 1]$ reflecting the assessed likelihood of anomalous content.

\begin{tcolorbox}[
  colback=gray!3!white,
  colframe=gray!60!black,
  title=Anomaly Classification System Prompt ($p_{\text{score}}$),
  fonttitle=\bfseries
]
\small
\texttt{You are an expert AI system specialized in video anomaly detection. Analyze scene descriptions and classify them as NORMAL or ANOMALY.}\\[0.5em]
\textbf{ANOMALY INDICATORS:}\\
\texttt{- Violence: fighting, assault, weapons, threatening behavior}\\
\texttt{- Explosions: fire, smoke, blast effects, debris, damaged structures}\\
\texttt{- Accidents: car collisions, crashes, overturned vehicles}\\
\texttt{- Theft: shoplifting, robbery, stealing, grabbing items and fleeing}\\
\texttt{- Medical emergencies: injuries, falls, people on ground needing help}\\
\texttt{- Suspicious behavior: inappropriate running, chasing, panic, fleeing}\\
\texttt{- Property damage: vandalism, destruction, breaking}\\
\texttt{- Abandoned objects: unattended bags, suspicious packages}\\
\texttt{- Crowd anomalies: panic, stampede, unusual mass movements}\\
\texttt{- Fire or smoke visible}\\
[0.5em]
\textbf{NORMAL INDICATORS:}\\
\texttt{- Routine activities: walking, standing, sitting, talking normally}\\
\texttt{- Shopping, browsing, work, recreational activities}\\
\texttt{- Vehicles in designated areas: roads, parking lots, bike lanes}\\
\texttt{- Regular traffic flow}\\
\texttt{- Appropriate social interactions}\\
\texttt{- quiet scenes with no damage or suspicious objects}
\texttt{-Respond ONLY with a JSON object containing the anomaly score:
{
  "anomaly score": 0.0-1.0
}}
\end{tcolorbox}

\subsection{Video-Level Temporal Captioning Prompt}

For frames that exceed the threshold $\tau_f$, we extract a temporal clip and process it through VideoLLaMA3-7B using the following prompt $p_{\text{vid}}$:

\begin{tcolorbox}[
  colback=gray!3!white,
  colframe=gray!60!black,
  title=Temporal Multi-Frame Captioning Prompt ($p_{\text{vid}}$),
  fonttitle=\bfseries
]
\small
\texttt{Describe the actions, movements, and events visible in this sequence.}\\[0.5em]
\texttt{Specifically identify any suspicious, unusual, or anomalous behavior such as: theft, shoplifting, violence, fighting, accidents, burglary, or other criminal activities.}
\end{tcolorbox}

This prompt is deliberately more concise than the frame-level captioning prompt, as the video model receives multiple frames providing temporal context.The emphasis on actions and movements leverages the video model's capacity to reason about dynamics across frames, complementing frame-level static analysis.

\subsection{Example Pipeline Outputs}

To illustrate the complete inference chain, we provide representative examples showing the progression from visual input through captioning to final anomaly scoring.

\vspace{0.5em}
\noindent\textbf{Example 1: Anomalous Scene (Explosion)}
\begin{tcolorbox}[
  colback=red!3!white,
  colframe=red!60!black,
  title=Delta-Triggered Frame (Sudden Event),
  fonttitle=\bfseries
]
\small
\textbf{Generated Caption:}\\
\textit{``The scene shows a large explosion occurring in an urban street. Thick black smoke and orange flames are visible rising from the center of the frame. Debris is scattered across the road. Several people are running away from the blast site in apparent panic. Damaged vehicles are visible near the explosion epicenter. The environmental context suggests a commercial district with storefronts visible in the background.''}\\[0.5em]
\textbf{Triggering Signal:} Delta ($\delta_t = 0.42$, threshold $\tau_{\delta,t} = 0.18$)\\[0.3em]
\textbf{Anomaly Score:} $s^{\text{frame}} = 0.95$\\[0.3em]
\textbf{Video-Level Gating:} $s^{\text{frame}} = 0.95 > \tau_f = 0.4$ $\rightarrow$ invoke video-level analysis\\[0.3em]
\textbf{Video-Level Score:} $s^{\text{vid}} = 0.91$\\[0.3em]
\textbf{Fusion Weight:} $\lambda_\delta = 0.6$ (delta-triggered $\rightarrow$ frame-level dominant)\\[0.3em]
\textbf{Fused Score:} $s^{\text{fused}} = \lambda_\delta \cdot s^{\text{frame}} + (1-\lambda_\delta) \cdot s^{\text{vid}} = 0.6 \times 0.95 + 0.4 \times 0.91 = 0.934$
\end{tcolorbox}

\vspace{0.5em}
\noindent\textbf{Example 2: Anomalous Scene (Static Anomaly)}
\begin{tcolorbox}[
  colback=red!3!white,
  colframe=red!60!black,
  title=Drift-Triggered Frame (Persistent Anomaly),
  fonttitle=\bfseries
]
\small
\textbf{Generated Caption:}\\
\textit{``A person is lying motionless on the ground in a pedestrian walkway. The individual appears to be in a prone position with no visible movement. Several pedestrians are walking past without stopping. An unattended bag is visible next to the person. The scene suggests a possible medical emergency or collapse.''}\\[0.5em]
\textbf{Triggering Signal:} Drift ($\rho_t = 0.31$, threshold $\tau_{\rho,t} = 0.12$)\\[0.3em]
\textbf{Anomaly Score:} $s^{\text{frame}} = 0.78$\\[0.3em]
\textbf{Video-Level Gating:} $s^{\text{frame}} = 0.78 > \tau_f = 0.4$ $\rightarrow$ invoke video-level analysis\\[0.3em]
\textbf{Video-Level Score:} $s^{\text{vid}} = 0.85$ (confirms persistent anomaly)\\[0.3em]
\textbf{Fusion Weight:} $\lambda_\rho = 0.4$ (drift-triggered $\rightarrow$ video-level dominant)\\[0.3em]
\textbf{Fused Score:} $s^{\text{fused}} = \lambda_\rho \cdot s^{\text{frame}} + (1-\lambda_\rho) \cdot s^{\text{vid}} = 0.4 \times 0.78 + 0.6 \times 0.85 = 0.822$
\end{tcolorbox}

\vspace{0.5em}
\noindent\textbf{Example 3: Normal Scene (Unselected Frame)}
\begin{tcolorbox}[
  colback=green!3!white,
  colframe=green!60!black,
  title=Unselected Frame (Below Signal Thresholds),
  fonttitle=\bfseries
]
\small
\textbf{Delta Signal:} $\delta_t = 0.05 < \tau_{\delta,t} = 0.15$ $\rightarrow$ not triggered\\[0.3em]
\textbf{Drift Signal:} $\rho_t = 0.08 < \tau_{\rho,t} = 0.11$ $\rightarrow$ not triggered\\[0.3em]
\textbf{Frame Selection:} $\mathcal{I}(t) = \mathcal{I}_\delta(t) \lor \mathcal{I}_\rho(t) = 0$ $\rightarrow$ frame not selected\\[0.3em]
\textbf{VLM Analysis:} Skipped (no captioning or scoring performed)\\[0.3em]
\textbf{Final Score:} $s_t = s_d = 0.3$ (default normality prior)
\end{tcolorbox}

These examples illustrate the complete SAAF pipeline: frames are selected only when delta or drift signals exceed their adaptive thresholds, video-level analysis is invoked only when the frame-level score exceeds $\tau_f = 0.4$, and unselected frames receive the default score $s_d$

\section{Limitations}

SAAF's adaptive thresholding assumes the warmup period contains representative normal content to establish baseline statistics. When anomalies begin immediately or dominate early frames, the EMA-based thresholds adapt to abnormal patterns as if they were normal, suppressing subsequent signal activations. This baseline contamination effect is evident in videos where anomalous events start at frame zero.

The dual-signal mechanism operates on stride-16 sampled frames, which limits temporal resolution for brief anomalies. On ShanghaiTech, where events typically span 40--60 frames, this coarse sampling provides only 2--4 signal observations per anomaly, reducing detection reliability compared to longer-duration events.

Finally, SAAF's accuracy ultimately depends on VLM reasoning capability. When anomalies exhibit subtle visual signatures, such as gradual fire progression in low-light conditions or fine-grained actions, correct frame selection cannot compensate for limitations in downstream semantic understanding. This coupling means that advances in VLM capabilities would directly benefit SAAF without architectural changes.

\section{Broader Impact}

SAAF's zero-shot design avoids encoding dataset-specific biases into learned parameters, though biases in underlying VLMs may still influence scoring. From an environmental perspective, SAAF's sampling reduces substantial computational cost compared to uniform processing, lowering the energy footprint of large-scale video analysis. As VLM deployment scales across infrastructure worldwide, such efficiency gains become increasingly consequential for sustainable AI.

\clearpage

\end{document}